\PassOptionsToPackage{table, xcdraw}{xcolor} % before any packages

\documentclass[fleqn,10pt]{wlscirep}
\usepackage[utf8]{inputenc}
\usepackage[T1]{fontenc}

\usepackage{xcolor} 

\usepackage{times}
\usepackage{epsfig}
\usepackage{graphicx}
\usepackage{amsmath,amsfonts,amssymb}
\usepackage{movie15}
\usepackage{mathptmx}  % Times Roman for text and math
\usepackage{newtxtext,newtxmath}  % Better Times-compatible math font

\usepackage{multirow}
\usepackage{threeparttable}
\usepackage{stackengine}
\usepackage{footnote}
\usepackage{array}
\usepackage{balance}
\usepackage{booktabs}
\usepackage{comment}
\usepackage{arydshln}
\usepackage{makecell}
\usepackage{algorithm}
\usepackage{algpseudocode}
\usepackage{caption}
\usepackage{subcaption}  % For subfigures
\usepackage{tikz}
\usepackage{pifont}
\usepackage{float}

\DeclareMathAlphabet{\mathbbold}{U}{bbold}{m}{n}

\makeatletter
\newcommand\figcaption{\def\@captype{figure}\caption}
\newcommand\tabcaption{\def\@captype{table}\caption}
\makeatother

\def\best#1{{\bf {#1}}}

\def\etal{\emph{et al.}}
\newcolumntype{P}[1]{>{\centering\arraybackslash}p{#1}}
\newcolumntype{C}[1]{>{\centering}m{#1}}

\makeatletter
\renewcommand{\toprule}{\specialrule{1.0pt}{0pt}{0pt}}    % Thick top rule
\renewcommand{\midrule}{\specialrule{0.5pt}{0pt}{0pt}}   % Medium middle rule
\renewcommand{\bottomrule}{\specialrule{1.0pt}{0pt}{0pt}} % Thick bottom rule
\makeatother

\newcommand{\improve}[1]{\scriptsize \textcolor{green}{$\blacktriangle$} #1}  % Green, left-pointing
\newcommand{\degrade}[1]{\scriptsize \textcolor{red}{$\blacktriangledown$} #1}   % Red, right-pointing

\newcolumntype{C}[1]{>{\centering\arraybackslash}m{#1}}  % Centered column with fixed width
\newcolumntype{|}{!{\vrule width 1pt}}                 % Vertical lines with 1pt thickness

\usepackage{arydshln}

\definecolor{newblue}{RGB}{107,153,208}
\definecolor{newyellow}{RGB}{246,195,66}

\usepackage{pifont}% http://ctan.org/pkg/pifont

\usepackage{hyperref}

\newcommand{\xg}[1]{\textcolor{violet}{[\textit{#1}]}}

\title{Sensing Cardiac Health Across Scenarios and Devices: A Multi-Modal Foundation Model Pretrained on Heterogeneous Data from 1.7 Million Individuals}

\empty

\author[1]{Xiao Gu}
\author[1,2,3]{Wei Tang}
\author[4]{Jinpei Han}
\author[1]{Veer Sangha}
\author[1]{Fenglin Liu}
\author[5]{Shreyank N Gowda}
\author[6]{Antonio~H.~Ribeiro}
\author[7]{Patrick~Schwab}
\author[7]{Kim~Branson}
\author[1,8]{Lei~Clifton}
\author[9]{Antonio~Luiz~P.~Ribeiro}
\author[1,10]{Zhangdaihong Liu}
\author[1,10]{David~A.~Clifton}
\affil[1]{Department of Engineering Science, University of Oxford, Oxford OX3 7DQ, UK}
\affil[2]{Department of Mathematics, City University of Hong Kong, Hong Kong}
\affil[3]{Hong Kong Center for Cerebro-Cardiovascular Health Engineering, Hong Kong}
\affil[4]{Brain and Behaviour Lab, Imperial College London, London SW7 2AZ, UK}
\affil[5]{School of Computer Science, University of Nottingham, Nottingham NG8 1BB, UK}
\affil[6]{Department of Information Technology, Uppsala University, Uppsala, Sweden}
\affil[7]{GlaxoSmithKline, London, UK}
\affil[8]{Nuffield Department of Primary Care Health Sciences, University of Oxford, Oxford OX2 6GG, UK}
\affil[9]{Department of Internal Medicine, Faculdade de Medicina, and Telehealth Center and Cardiology Service, Hospital das Clínicas, Universidade Federal de Minas Gerais, Belo Horizonte, Brazil}
\affil[10]{Oxford Suzhou Centre for Advanced Research, University of Oxford, Suzhou 215123, China}

\begin{abstract}
 
Cardiac biosignals, such as electrocardiograms (ECG) and photoplethysmograms (PPG), are of paramount importance for the diagnosis, prevention, and management of cardiovascular diseases, and have been extensively used in a variety of clinical tasks. Conventional deep learning approaches for analysing these signals typically rely on homogeneous datasets and static bespoke models, limiting their robustness and generalizability across diverse clinical settings and acquisition protocols. In this study, we present a cardiac sensing foundation model (CSFM) that leverages advanced transformer architectures and a generative, masked pretraining strategy to learn unified representations from vast, heterogeneous health records. Our model is pretrained on an innovative multi-modal integration of data from multiple large-scale datasets (including MIMIC-III-WDB, MIMIC-IV-ECG, and CODE), comprising cardiac signals and the corresponding clinical or machine-generated text reports from approximately 1.7 million individuals. We demonstrate that the embeddings derived from our CSFM not only serve as effective feature extractors across diverse cardiac sensing scenarios, but also enable seamless transfer learning across varying input configurations and sensor modalities. Extensive evaluations across diagnostic tasks, demographic information recognition, vital sign measurement, clinical outcome prediction, and ECG question answering reveal that CSFM consistently outperforms traditional one-modal-one-task approaches. Notably, CSFM exhibits robust performance across multiple ECG lead configurations from standard 12-lead systems to single-lead setups, and in scenarios where only ECG, only PPG, or a combination thereof is available. These findings highlight the potential of CSFM as a versatile and scalable solution for comprehensive cardiac monitoring across both resource-rich and resource-constrained healthcare environments.

\end{abstract}
\begin{document}

\flushbottom
\maketitle

\thispagestyle{empty}

\section*{Introduction}

%% importance of cardiac sensing from hospital to home
Cardiovascular diseases are among the leading causes of morbidity and mortality worldwide, underscoring the need for accurate and timely diagnostic methods \cite{kaptoge2019world}. In clinical practice, cardiac biosignals—most notably electrocardiograms (ECGs) and photoplethysmograms (PPGs)—serve as critical tools for diagnosing, preventing, and managing these conditions \cite{bayoumy2021smart}. ECGs capture the electrical impulses generated by the heart, providing essential information on cardiac rhythm and conduction pathways. On the other hand, PPGs track fluctuations in blood volume through optical sensors, enabling non-invasive monitoring of peripheral blood flow and cardiac output. The synergistic integration of these biosignals holds significant potential for digital health innovation, with applications ranging from acute clinical settings \cite{sundrani2023predicting} to continuous home-based care \cite{steinhubl2018effect}.
\begin{figure}[!ptb]
    \centering
    \includegraphics[width=0.9\linewidth]{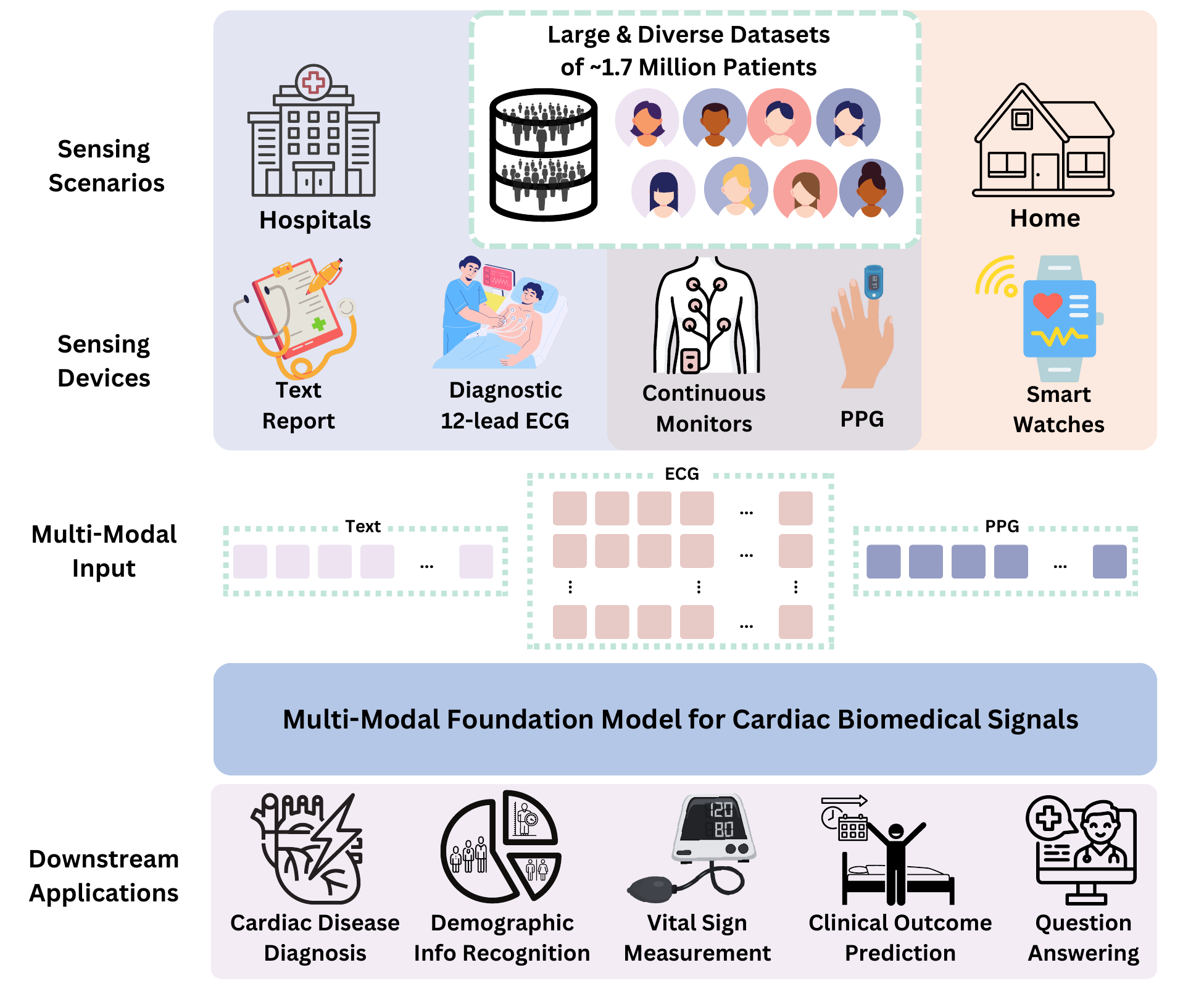}
    \caption{\textbf{Illustration of the cardiac sensing foundation model (CSFM), capable of taking heterogeneous biomedical signals as input and versatile for different cardiac health-related downstream tasks.} Given the diversity in sensing scenarios and devices, the collected biomedical sensing data are varied in both the signal types and the channels for each record. We trained CSFM on an innovative integration of multiple cardiac sensing datasets collected from around 1.7 million individuals and assessed its performance in diverse healthcare scenarios. The generalization capability of CSFM is tested across different scenarios from hospital to home, on representative tasks. These include cardiac disease diagnosis, demographic information reocgnition, vital sign measurement, clinical outcome prediction (spanning short-term ICU alerts to long-term mortality), and ECG-based question answering. }  
    \label{fig:figure1}
\end{figure}

Advances in sensing technologies have dramatically reshaped the acquisition landscape of cardiac biosignals \cite{gu2023beyond}, as illustrated in \figureautorefname~\ref{fig:figure1}. In standard hospital settings, comprehensive 12-lead ECGs are routinely used to capture detailed cardiac activity. These recordings are usually supplemented with contextual clinical annotations—either from cardiologists or generated automatically by software—to enhance their diagnostic and prognostic value. In intensive care units (ICUs), a more streamlined approach is adopted, with fewer-lead ECGs often paired with PPG signals to facilitate real-time monitoring and early detection of adverse events. Additionally, in ICU step-down wards as well as in-home and community settings, wearable devices such as smart wrist-worn sensors or patches are increasingly employed to capture ECG or PPG signals for continuous monitoring.

%\xg{Importance of texts}

% challenges

% task itself
 \begin{figure}[tbp]
    \centering
    \begin{subfigure}[t]{\linewidth}
        \makebox[0pt][l]{\raisebox{6.cm}{\hspace{-1.cm}\Large{\textbf{a}}}}
        \centering
        \includegraphics[width=0.65\linewidth]{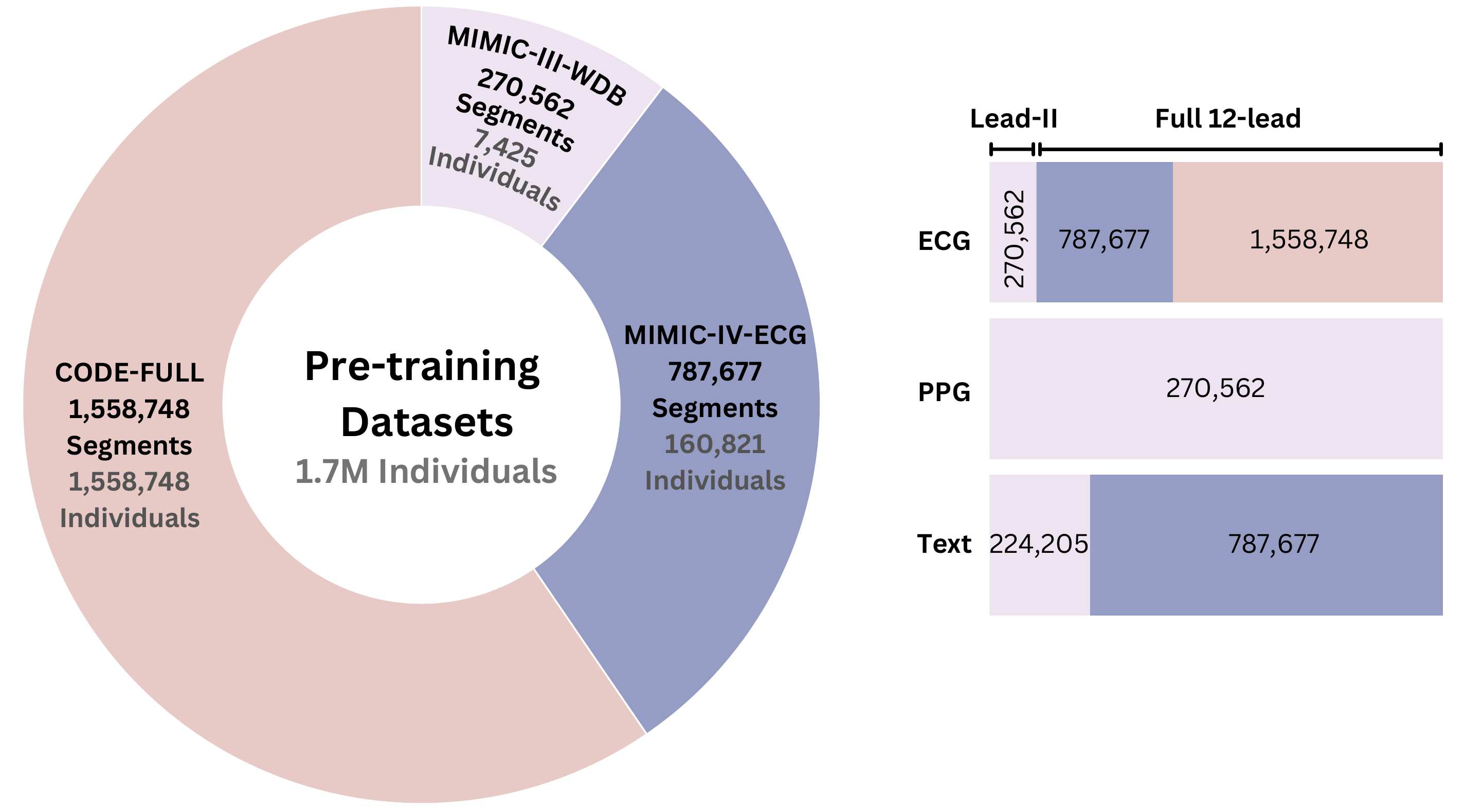}
        \label{fig:pretrain_data}
    \end{subfigure}
     
    \begin{subfigure}[t]{0.99\linewidth}
        \makebox[0pt][l]{\raisebox{7.cm}{\hspace{0.cm}\Large{\textbf{b}}}}
        \includegraphics[width=\linewidth]{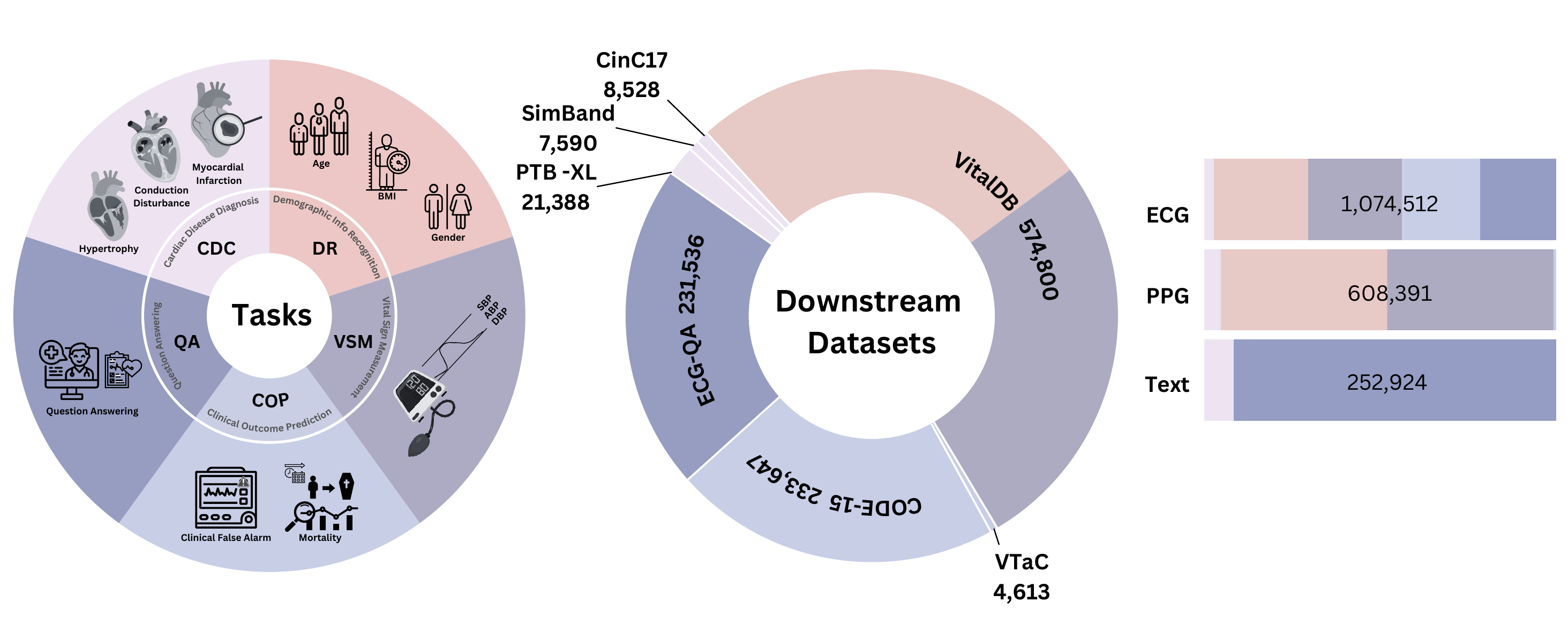}
        \label{fig:downstream_data}
    \end{subfigure}

    \caption{\textbf{Statistics of our training and testing datasets.} \textbf{a.} Illustration of pretraining datasets. Our pretraining dataset is aggregated from heterogeneous records across multiple sources, including MIMIC-III-WDB, MIMIC-IV-ECG, and CODE-Full. It is also noteworthy that while MIMIC-IV and MIMIC-III-WDB may contain overlapping subjects, all records are de-identified, making subject linkage impossible. Their data segments were collected from distinct clinical scenarios. The left plot illustrates the number of recorded segments across datasets, while the right plot represents the number of segments across different signal modalities. \textbf{b.} Illustration of downstream tasks and datasets. Our downstream evaluation spans five cardiology-related scenarios, including cardiovascular disease diagnosis (CDD), demographic information recognition (DIR), Vital Sign Measurement (VSM), Clinical Outcome Prediction (COP), and Question Answering (QA). The downstream datasets were collected from multiple sources, including CinC17 \cite{clifford2017af}, PTB-XL \cite{wagner2020ptb}, SimBand \cite{shashikumar2017deep}, VTaC \cite{lehman2024vtac}, CODE-15 \cite{lima2021deep,ribeiro2020automatic}. The figure on the right summarizes the distribution of signals across different modalities.}
    \label{fig:data}
\end{figure}

However, traditional analytical methods often lack the scalability to coordinate data acquired from such diverse devices and environments. Conventional approaches typically require bespoke models tailored to specific signal types, sensing modalities, and clinical tasks. These models are frequently developed from scratch and rely on large-scale, annotated datasets with consistent data formats (e.g., identical ECG channel configurations or uniform signal types). In clinical practice, such large-scale datasets are often unavailable due to the inherent heterogeneity of cardiac biosignals and the specialized expertise required for accurate annotation. Consequently, these methods often yield suboptimal performance on a single limited dataset, given the lack of access to sufficiently comprehensive and uniform data.

% data fragmented
Furthermore, models developed on fragmented datasets often exhibit limited transferability and may not be directly applicable across diverse healthcare environments. For instance, a model trained on data from routinely collected 12-lead ECGs may fail to generalize to settings in low- and middle-income countries (LMICs), where portable or wearable devices (e.g., wearable ECG/PPG) provide more affordable solutions. Traditional methods, e.g., those based on convolutional networks, are typically channel-dependent \cite{sangha2022automated,hannun2019cardiologist}, necessitating modifications to the network architecture to accommodate varying channel configurations. This disparity not only highlights significant inequities in access to state-of-the-art analytical tools but also underscores the urgent need for versatile, scalable solutions capable of robust performance across diverse clinical contexts.

% motivation

On the other hand, in the field of deep learning, there is an emerging trend toward developing foundation models that can derive generic representations through self-supervised training on large-scale datasets \cite{bommasani2021opportunities}. Remarkable achievements have been realized in both natural language processing \cite{achiam2023gpt,touvron2023llama} and computer vision \cite{kirillov2023segment, dosovitskiy2020vit}. However, in the realm of cardiac biosignals, existing foundation models are predominantly based on ECG data and are largely confined to standard 12-lead configurations \cite{vaid2023foundational}. This restriction to consistent data dimensionalities significantly limits their utility in broader clinical contexts, where diverse sensing modalities and heterogeneous data formats are common.

% contribution and highlights
To address these challenges, we develop a foundation model, the cardiac sensing foundation model (CSFM; \figureautorefname{}~\ref{fig:figure1}), by incorporating heterogeneous data types—including ECGs from various clinical settings, PPG signals, and accompanying clinical annotations—to enable robust, scalable performance across diverse healthcare environments. We train our model using advanced transformer architectures, originally developed for natural language processing and renowned for their ability to process sequential data and capture intricate dependencies \cite{vaswani2017attention,xu2023multimodal}. This sequential processing capability enables CSFM to effectively manage and integrate multi-modal, multi-channel information, making it particularly well-suited for analyzing cardiac biosignals. We employ masked training strategies, where signals are partially obscured across temporal and channel dimensions during pretraining, to facilitate pretraining on heterogeneous data inputs, which is a critical approach for aggregating cardiac sensing related data from diverse sources.

The CSFM is pretrained on an innovative integration of cardiac biosignals and associated cardiologist descriptions collected from approximately 1.7 million individuals. We systematically evaluate this framework on datasets gathered from different scenarios and devices, demonstrating outstanding performance in varied scenarios, including demographic information analysis, cardiovascular disease classification, vital sign measurement, clinical outcome prediction, and ECG-based question answering. Unlike traditional biosignal analysis models that are typically specialized for specific tasks or data types, CSFM can learn generalized representations and adapt to a wide range of downstream applications, offering a versatile and scalable tool for comprehensive cardiac biosignal analysis.

\section*{Results}

\subsection*{Pretraining on vast and heterogeneous cardiac health records}
The cardiac sensing foundation model leverages a generative pretraining approach, masked modeling, to learn generic representations from diverse biomedical signals. The pretraining process utilized data from multiple large-scale datasets, including MIMIC-III-WDB (waveform databases) \cite{johnson2016mimic}, MIMIC-IV-ECG \cite{gow2023mimic}, and a privately-held large-scale CODE dataset \cite{lu2024decoding, ribeiro2020automatic}, as shown in \figureautorefname{}~\ref{fig:data}a. To enhance the diversity and comprehensiveness of our pretraining data, we integrated cardiac biosignals with both machine-generated and clinical notes where available. Specifically, we linked the MIMIC-III-WDB ECG/PPG data with corresponding ECG reports extracted from the MIMIC-III clinical database \cite{johnson2016mimic}, and for MIMIC-IV-ECG, we associated ECG signals with machine-generated reports. Further details related to the statistics of the pretraining data are available in \figureautorefname{}~\ref{fig:data}a, as well as the Supplementary Material Section S1. Based on this integration, we applied a masking strategy to obscure channel-wise information and temporal-wise information during pretraining, enabling the model to generalize effectively across varied input configurations. To accommodate different computational and deployment needs, we developed three versions of CSFM: CSFM-Tiny, CSFM-Base and CSFM-Large, corresponding to tiny, base, and large in terms of parameter count, respectively.

\begin{figure*}[!htbp]
  \centering
%%%%%%%%%%%%%%%%%%%%%%%%%%%%%%%%%%%%%%%%%%%%%%%%%%%%%%%%%%%%%%%%%%%%%%%%%%%%%%%
  % First item: Cardiovascular Disease Performance
  \begin{minipage}{0.95\linewidth}
    \centering
    \makebox[0pt][l]{\raisebox{0cm}{\hspace{-8.8cm}\Large{\textbf{a}}}}
    \includegraphics[width=\linewidth]{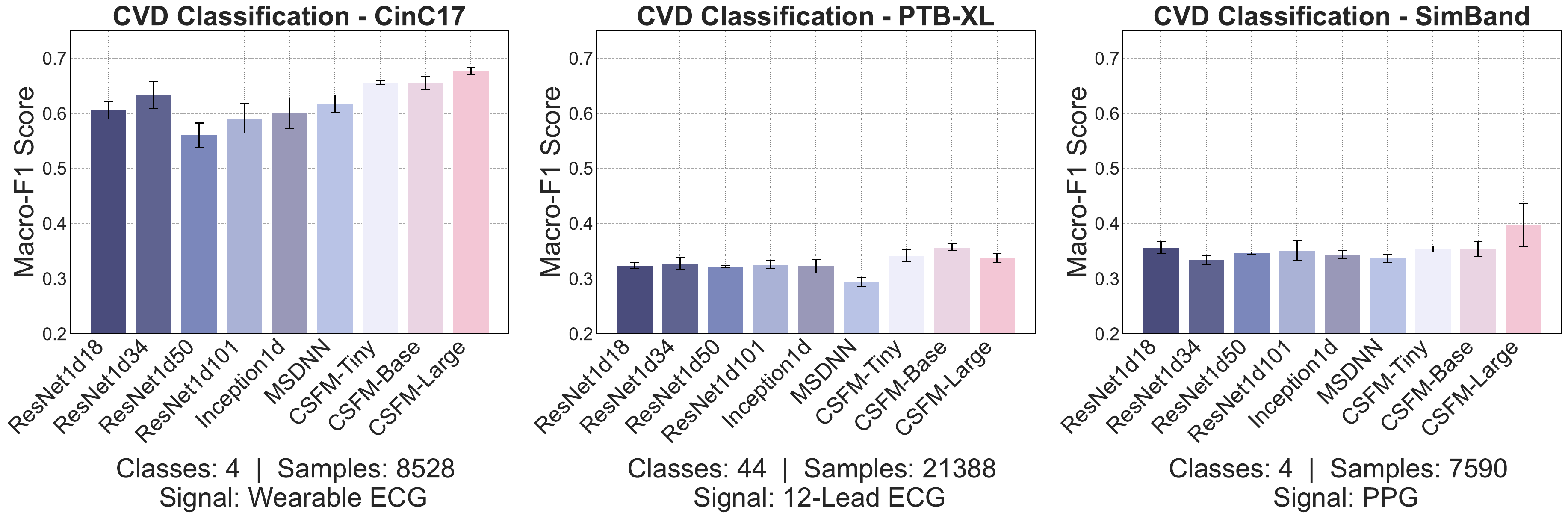}
    \vspace{-15pt}
    % \subcaption{}
    \label{fig:cvd}
  \end{minipage}

  % \vspace{1em}
  %%%%%%%%%%%%%%%%%%%%%%%%%%%%%%%%%%%%%%%%%%%%%%%%%%%%%%%%%%%%%%%%%%%%%%%%%%%%%%%
  % Second item: VitalDB Demo
  \begin{minipage}{0.95\linewidth}
    \centering
    \makebox[0pt][l]{\raisebox{-0.2cm}{\hspace{-8.8cm}\Large{\textbf{b}}}}
    \includegraphics[width=\linewidth]{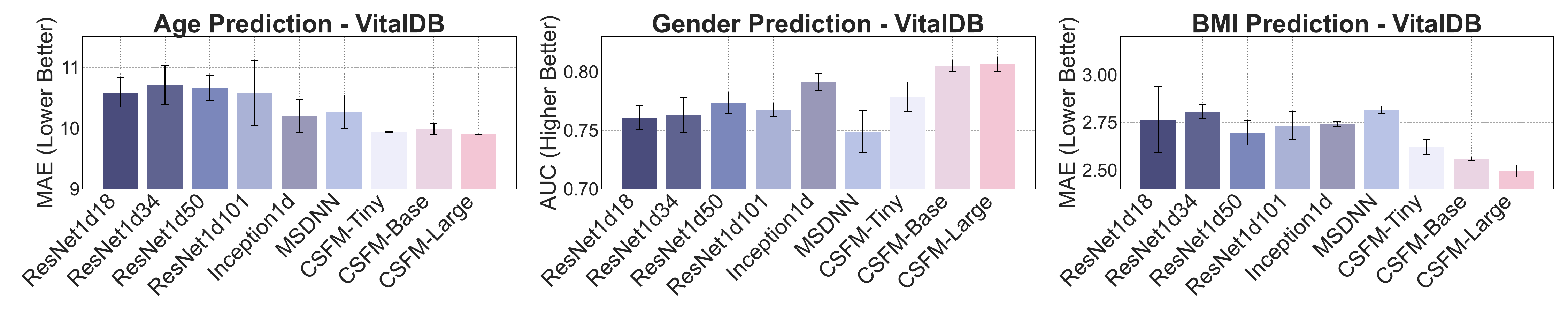}
    \vspace{-15pt}
    % \subcaption{}
    \label{fig:vital_demo}
  \end{minipage}

  % \vspace{1em}
  %%%%%%%%%%%%%%%%%%%%%%%%%%%%%%%%%%%%%%%%%%%%%%%%%%%%%%%%%%%%%%%%%%%%%%%%%%%%%%%
  % Third item: Blood Pressure Waveform Reconstruction (two images side by side)
  \begin{minipage}{0.95\linewidth}
    \centering
    \makebox[0pt][l]{\raisebox{-0.cm}{\hspace{-8.8cm}\Large{\textbf{c}}}}
    \includegraphics[width=0.4\linewidth]{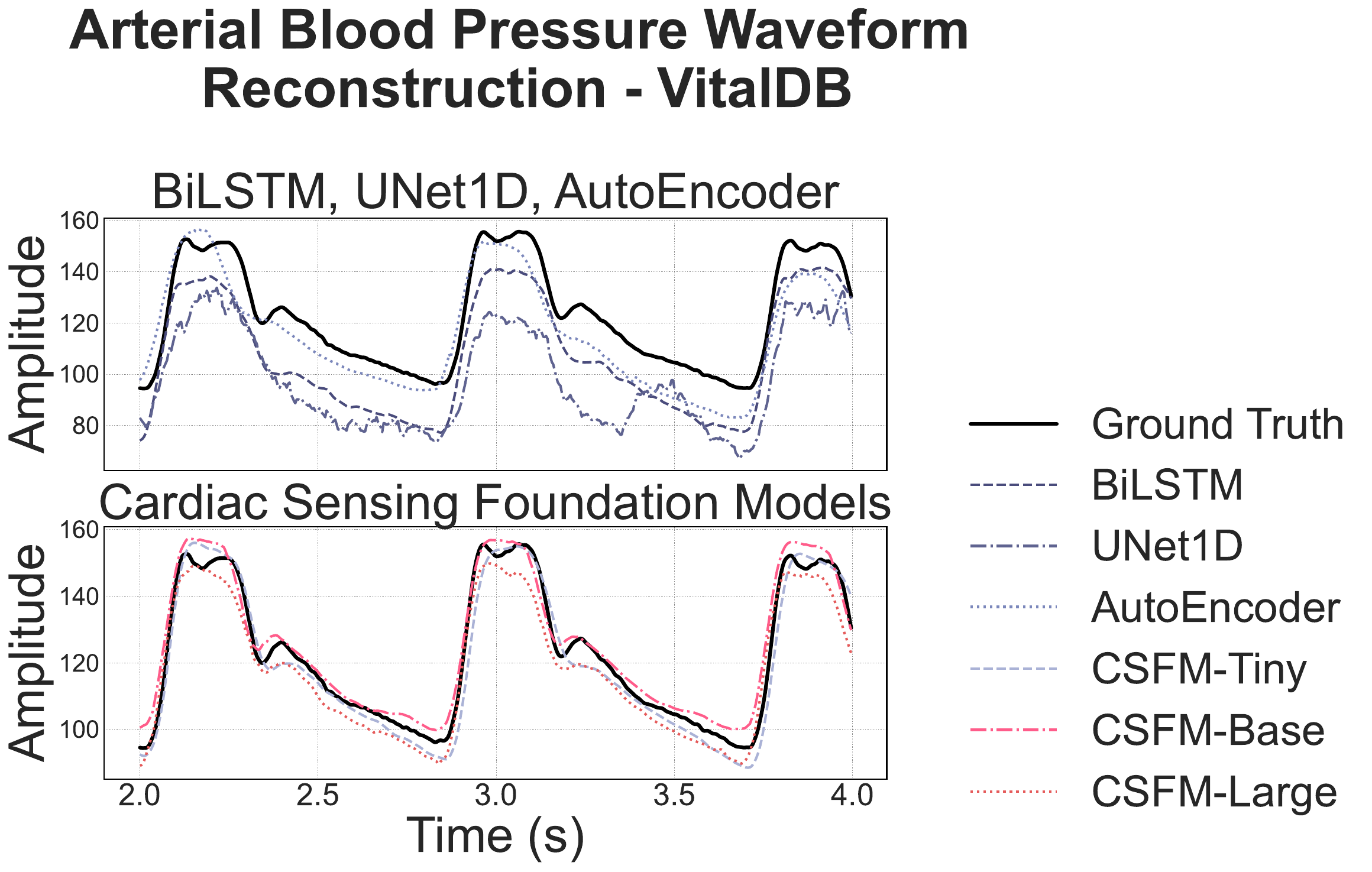}%
    \includegraphics[width=0.62\linewidth]{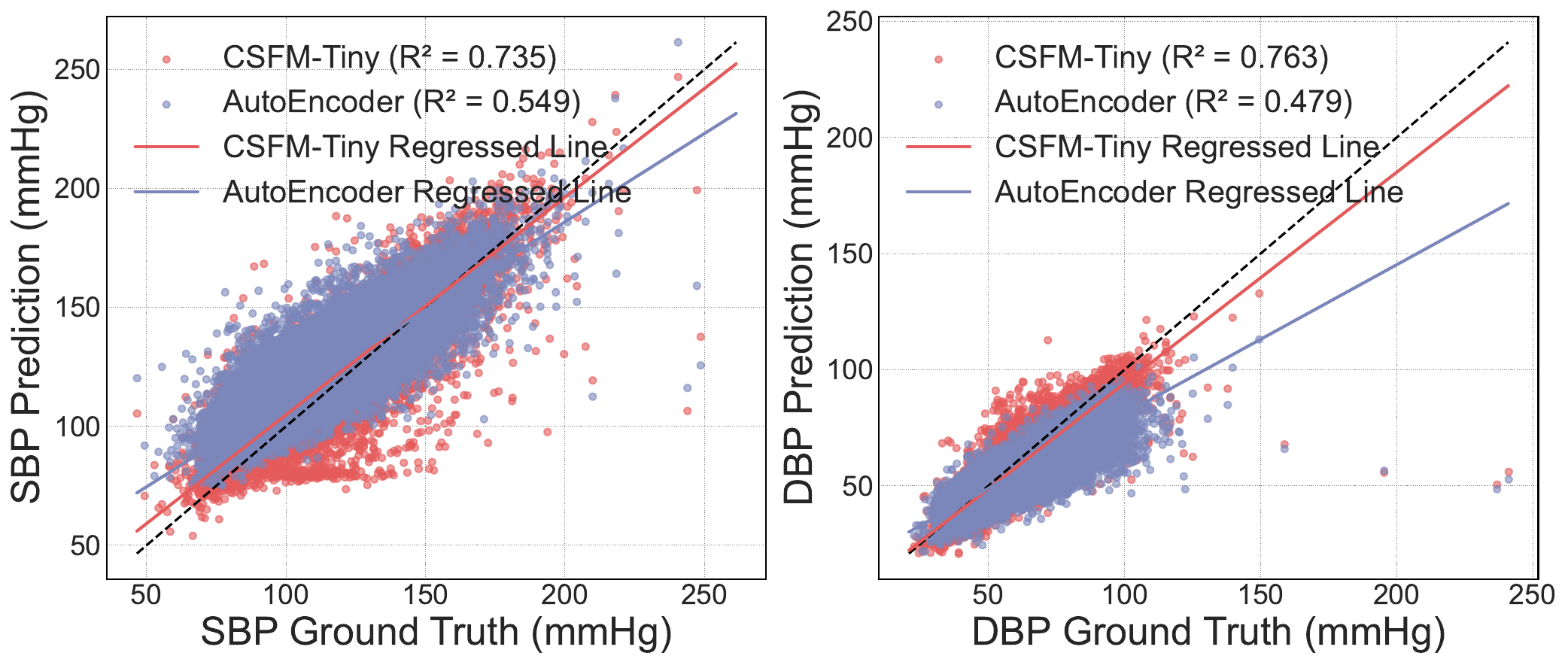}
    % \subcaption{}
    \label{fig:vital_bp}
    \vspace{-15pt}
  \end{minipage}

  % \vspace{1em}
  
  %%%%%%%%%%%%%%%%%%%%%%%%%%%%%%%%%%%%%%%%%%%%%%%%%%%%%%%%%%%%%%%%%%%%%%%%%%%%%%%
  % Fourth item: ROC Plots (two images side by side)
% Third row: Last three items on the same line (three columns)
  % Column 1: Blood Pressure Waveform Reconstruction (two images arranged vertically)
  \hfill
  % Column 2: ROC Plots (two images arranged side by side inside this column)
\centering
    \begin{minipage}[b]{0.33\linewidth}
      \vspace{0pt} % ensures proper baseline handling
      \centering
    \makebox[0pt][l]{\raisebox{0cm}{\hspace{-3cm}\Large{\textbf{d}}}}
      \includegraphics[width=\linewidth]{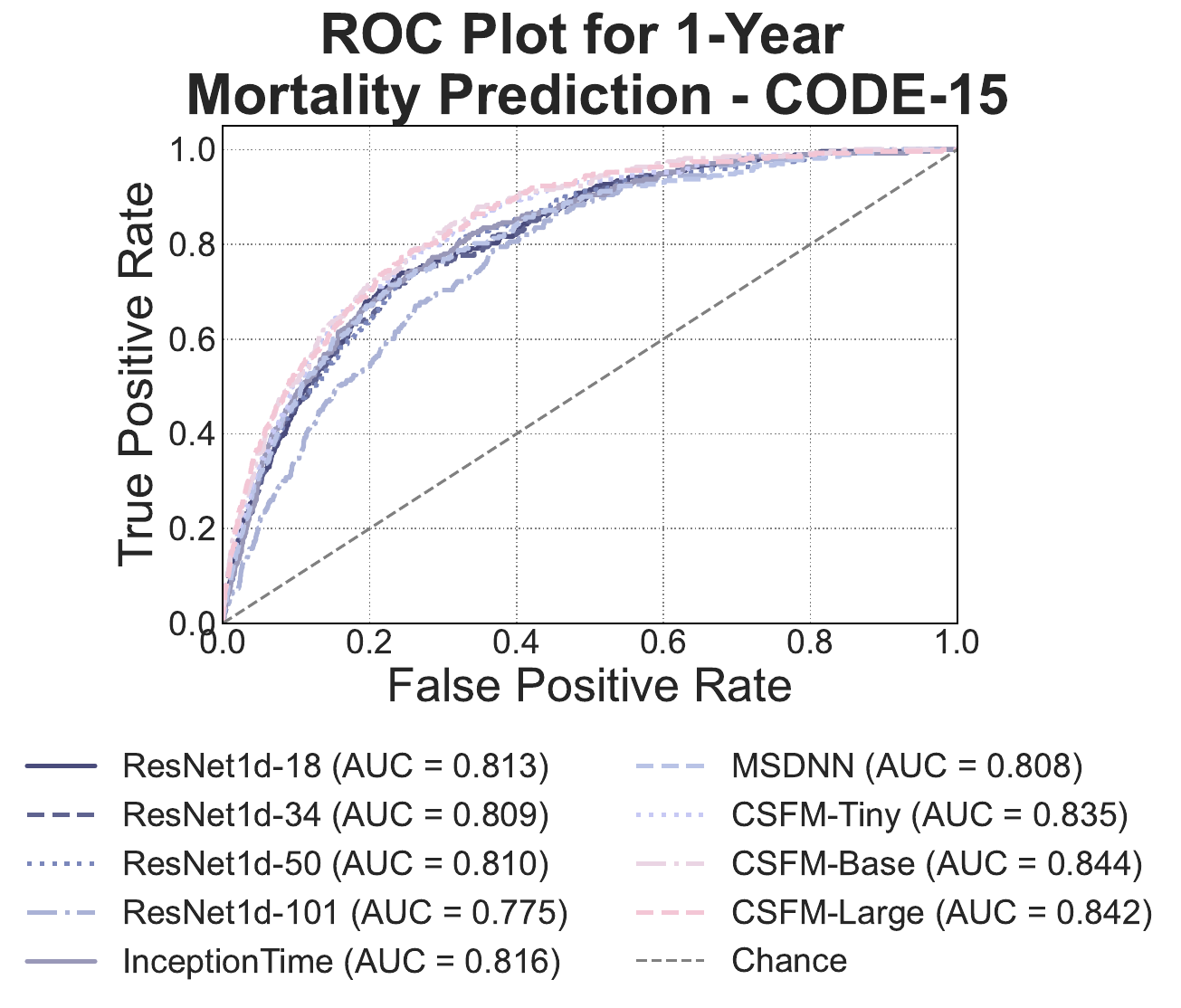}
      \vspace{-20pt}
      % \subcaption{1-Year mortality prediction based on 12-Lead diagnostic ECGs.}
      \label{fig:mortality}
    \end{minipage}
    \begin{minipage}[b]{0.33\linewidth}
      \vspace{0pt}
      \centering
    \makebox[0pt][l]{\raisebox{0cm}{\hspace{-3cm}\Large{\textbf{e}}}}
      \includegraphics[width=\linewidth]{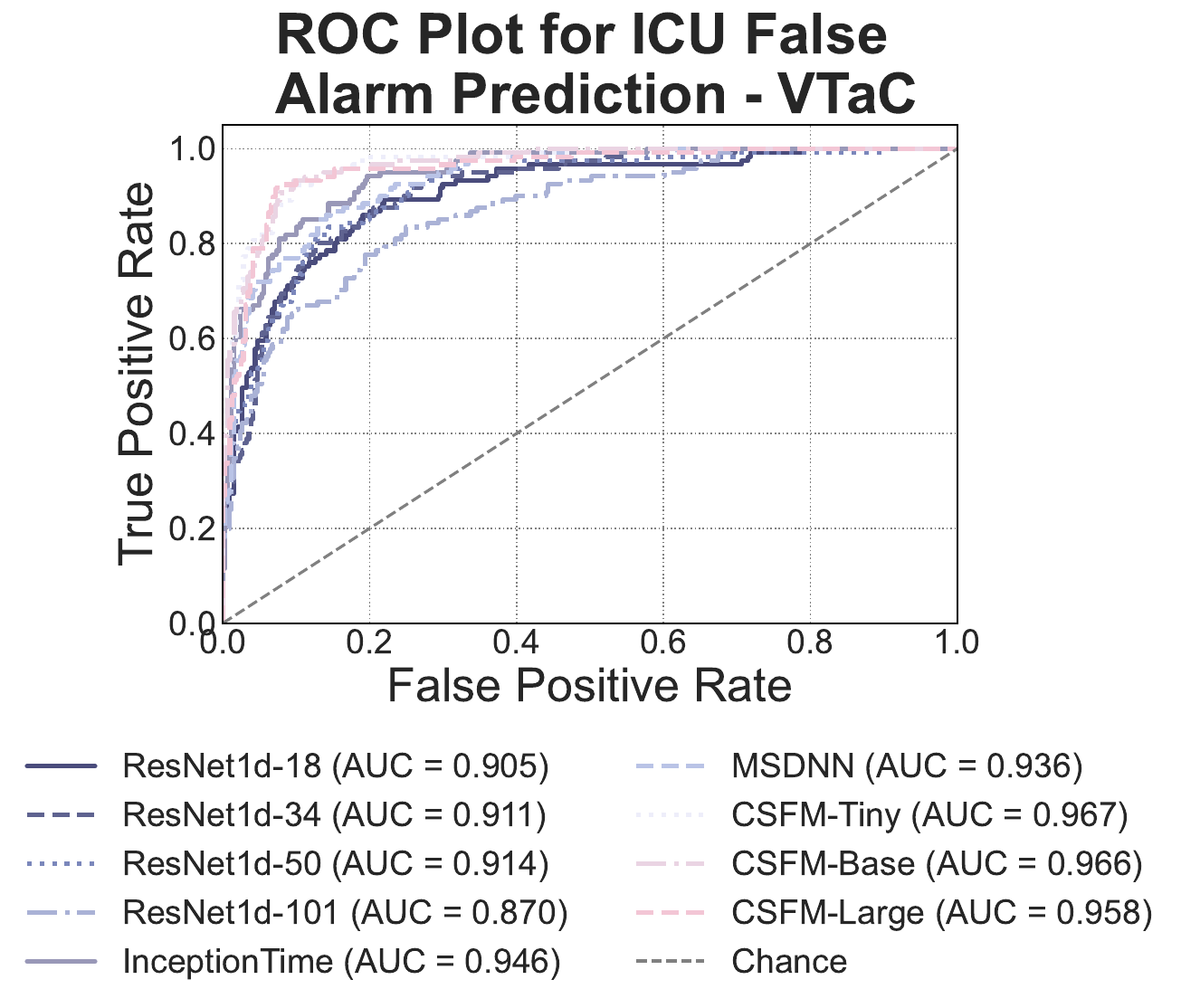}
      \vspace{-20pt}
      % \subcaption{ICU false alarm Prediction based on signals (ECG \& PPG) right before the alarm.} 
      \label{fig:icu}
    \end{minipage}
    \hfill
    \begin{minipage}[b]{0.32\linewidth}
      % \vspace{50pt}
      \centering
        % \subcaption{}
      % Table caption at the top:
      % \vspace{0.5em} % adjust spacing as needed between caption and table
      \makebox[0pt][l]{\raisebox{4.5cm}{\hspace{-0.5cm}\Large{\textbf{f}}}}
      \includegraphics[width=0.9\linewidth]{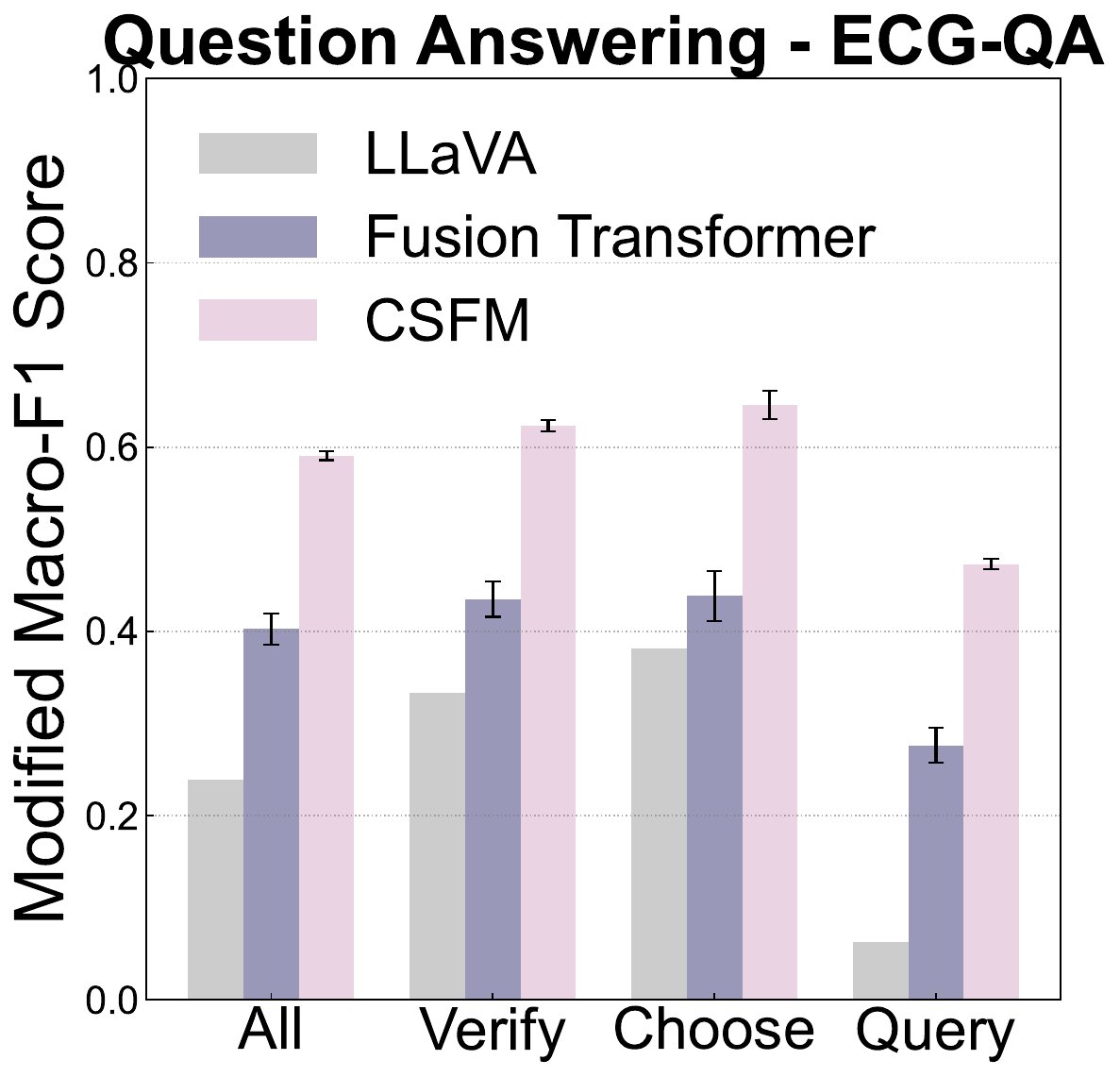}
      % \resizebox{\linewidth}{!}{
      % \begin{tabular}{lccc}
      %   \toprule
      %     \textbf{Methods} & \textbf{Verify} & \textbf{Choose} & \textbf{Query} \\
      %   \midrule
      %   % "#EEEEFA", "#EAD4E3", "#F3C6D5",  

      %  % M3AE \cite{NEURIPS2023_d0b67349}    & 0.761 & 0.850 & 0.836 \\ 
      %  %  MedViLL \cite{NEURIPS2023_d0b67349}   & 0.768 & 0.839 & 0.831 \\ 
      %  % Fusion Trans. \cite{NEURIPS2023_d0b67349}    & 0.725 & 0.797 & 0.791 \\
      %   \arrayrulecolor[HTML]{AAB2D6}
      %   \midrule       
      %   Majority & \\
      %   Fusion Trans \cite{NEURIPS2023_d0b67349}  & 0.711 &  0.793 & 0.781 \\ 
      %   % LLaVa3  & - & - & - \\
      %   \midrule       
      %   \arrayrulecolor{black}
      %   \midrule 
      %  \rowcolor[HTML]{EEEEFA} CSFM-Tiny    & 0.766 & 0.841 & 0.818 \\ 
      %  \rowcolor[HTML]{EAD4E3} CSFM-Base     & 0.771 & 0.860 & 0.838 \\ 
      %   \rowcolor[HTML]{F3C6D5} CSFM-Large     & 0.774 & 0.854 & 0.840 \\ 
      %   \bottomrule
      % \end{tabular}}
      % \label{tab:ecgqa}
    \end{minipage}
    % \vspace{-10pt}
  \hfill 
  \caption{\textbf{Overall performance across different healthcare scenarios, validated on corresponding downstream datasets, separately.} \textbf{a.} Cardiovascular disease diagnosis across different datasets. The performance was measured by Macro-F1 in terms of multi-label/class classification. \textbf{b.} Demographic information recognition. Age and BMI prediction (univariate regression) was measured by mean absolute error (MAE, lower is better), whereas gender prediction (binary classification) was measured by AUC (higher is better). \textbf{c.} Blood pressure waveform reconstruction based on Lead-II ECG and PPG as input. We compared both the error of derived numeric values (i.e., SBP and DBP), as well as the raw waveforms. The R-squared values of the derived SBP and DBP against the ground truths, were calculated. \textbf{d.} 1-Year mortality prediction based on 12-Lead diagnostic ECGs. Receiver operating characteristic (ROC) curve is presented. \textbf{e.} ICU false alarm prediction based on signals (ECG \& PPG) right before the alarm. Receiver operating characteristic (ROC) curve is presented. \textbf{f.} ECG Question Answering with paired ECGs and questions. Question answering was formulated as a multi-choice QA system in which, for each question template, the model selects the most appropriate answers from a set of candidate responses. Performance was measured using the macro-F1 score, computed over only the valid candidate answers for each question.}
  \label{fig:grouped}
\end{figure*}

\subsection*{Downstream evaluation across diverse cardiac sensing scenarios and devices}
The objective of cardiac sensing foundation model is to achieve robust generalization and exceptional flexibility across diverse sensing devices and medical scenarios, enabling seamless integration and practical application in real-world healthcare settings. 

The selected downstream tasks address various clinical applications of physiological waveforms, encompassing a wide range of healthcare needs. These tasks include demographic information analysis, cardiovascular disease classification (PTB-XL \cite{wagner2020ptb}, CinC17 \cite{clifford2017af}, SimBand \cite{shashikumar2017deep}), vital sign measurement (VitalDB \cite{lee2022vitaldb}), clinical outcome prediction (VTaC \cite{lehman2024vtac}, CODE-15 \cite{lima2021deep}), as well as question answering (ECG-QA). Among these, demographic information recognition, such as gender, BMI, and age, to uncover basic biological information encoded in cardiac biosignals. Cardiovascular disease classification supports the diagnosis of cardiac conditions by leveraging the rich information in the waveforms. Vital sign measurement enables the continuous monitoring of key physiological parameters (e.g., blood pressure), while clinical outcome prediction aids in risk stratification and long-term patient management. Additionally, question-answering tasks integrate both cardiac waveforms and associated textual data to provide interpretive insights and enhance decision-making. 

Among these tasks, vital sign measurement was considered as a dense regression task, where convolution layers are further added on top of CSFM to perform dense prediction, similar to Ranftl \etal{}\cite{ranftl2021vision}. For all other tasks, we added an additional fully connected layer to perform classification or univariate regression tasks.

The datasets used for downstream evaluation are collected from diverse real-world scenarios, ensuring comprehensive evaluation of CSFM’s adaptability and generalizability, with statistics shown in \figureautorefname{}~\ref{fig:data}b and further details in Supplementary Material Section S1. 

\subsection*{CSFM generalizes across different healthcare tasks}
\label{sec:task}
Our cardiac foundation model is highly adaptable across different tasks for specific healthcare scenarios. \figureautorefname{}~\ref{fig:grouped} compared the performance of CSFM with that of multiple basic/advanced deep learning models for medical times series (especially ECG) trained from scratch. These include classification/regression based models (ResNet1d-18/34/50/101, Inception1D \cite{ismail2020inceptiontime}, a Multi-Scale Deep Neural Network - MSDNN~\cite{lai2023practical}) and dense sequence to sequence regression models (BiLSTM, UNet1D, CNN based Autoencoder~\cite{gu2020cross}). They were tested across the aforementioned five downstream tasks. %and the overall performance are shown in \figureautorefname{}~\ref{fig:grouped}. 

To highlight the flexibility and generalizability of CSFM, we compared its fine-tuning performance with that of those compared models trained from scratch. This setup reflects the practical challenges of adopting ECG/PPG models in real-world applications. Traditional biomedical signal models are typically designed and deployed with fixed input dimensionalities and prediction tasks, making it intractable to direct transfer across scenarios, without architectural modifications. In contrast, CSFM enables seamless adaptation across diverse clinical settings, highlighting its potential as a versatile tool for comprehensive biosignal analysis.

\noindent\textbf{Cardiovascular Disease Diagnosis (Wearable ECG, PPG, 12-Lead ECG).} The performance of CSFM was evaluated on three datasets representing distinct sensing modalities and acquisition channels: CinC17 \cite{clifford2017af} (4 classes, multi-class classification), PTB-XL \cite{wagner2020ptb} (44 classes, multi-label classification), and SimBand \cite{shashikumar2017deep} (4 classes, multi-class classification). Each dataset was split subject-wise (80\% training, 10\% validation, 10\% testing). The best performance (mean across seeds) of our foundation model compared to traditional methods in each scenario is as follows: on CinC17, CSFM achieved a macro-F1 of 0.634 (95\% confidence interval (CI): [0.558, 0.710]) versus 0.677 (95\% CI: [0.656, 0.699]); on PTB-XL, it obtained 0.328 (95\% CI: [0.296, 0.361]) versus 0.357 (95\% CI: [0.338, 0.377]); and on SimBand, it reached 0.357 (95\% CI: [0.324, 0.391]) versus 0.398 (95\% CI: [0.279, 0.516]). These results, reported in terms of macro-F1, are shown in \figureautorefname{}~\ref{fig:grouped}a. In most cases, our CSFM model series substantially outperforms conventional learning strategies. Notably, on the SimBand dataset, although our CSFM-Large model achieves the best mean macro-F1 performance, its performance appears slightly more variable compared to other CSFM series. 

\noindent\textbf{Demographic Information Recognition (ECG, PPG).} This was evaluated on VitalDB dataset, by splitting data subject wise (80\% training, 10\% validation, 10\% testing). We analyzed the model performance in terms of Age (measured by meas absolute error (MAE), the lower the better), BMI (also measured by MAE), and Gender (measured by AUC, the higher the better). We train the model for these three tasks, separately. Our results indicate that CSFM consistently outperforms models trained from scratch across all metrics, as presented in \figureautorefname{}~\ref{fig:grouped}b.

\noindent\textbf{Vital Sign Measurement (ECG, PPG).} 
We utilized the VitalDB dataset to evaluate blood pressure measurement performance under calibration-based settings, as standardized in the original paper \cite{wang2023pulsedb}. We treats measurement predictions as a sequence-to-sequence regression problem. Specifically, our cardiac sensing foundation model first converts continuous ECG and/or PPG signals into continuous arterial blood pressure signals, and subsequently extracts the systolic (maximum, SBP) and diastolic (minimum, DBP) blood pressure values for comparison against ground truth. CSFM's performance was assessed in both stages, by comparing the waveform reconstruction quality as well as the numeric SBP and DBP values, as illustrated in \figureautorefname{}~\ref{fig:grouped}c with R-squared value calculated. Further details including MAE and root mean square error (RMSE) can be found in \tableautorefname{}~\ref{tab:bp_table} of the following section.

\noindent\textbf{Clinical Outcome Prediction (ECG, PPG).} 
In this scenario, we evaluated the predictive performance of our models by forecasting the likelihood of adverse events. Specifically, we performed analysis in two distinct settings: first, in acute ICU environments where it was used to identify false ICU alarms based on preceding ECG and PPG signals\cite{clifford2015physionet,lehman2024vtac} (as shown in \figureautorefname{}~\ref{fig:grouped}d, and second, for 1-year mortality prediction using standard 12-lead ECGs (as shown in \figureautorefname{}~\ref{fig:grouped}e. These evaluations demonstrate the versatility of our approach across both immediate critical care and long-term risk stratification tasks. The receiver operating characteristic (ROC) curves, which illustrate the accuracy of our predictions, are presented in \figureautorefname{}~\ref{fig:grouped}d,e.
It should be noted that CODE-15 is public small version of CODE-Full \cite{lima2021deep}, and in experimental settings we ensured that no training subjects in CODE-Full is available in validation/testing subset of CODE-15. Our results show that the AUC for the CSFM series reaches up to 0.844 for 1-year mortality prediction, compared to 0.816 for conventional deep learning methods trained from scratch. Additionally, the false alarm prediction task achieves superior performance relative to traditional approaches.

\noindent\textbf{ECG Question Answering (Text, ECG).}
We leveraged recently released ECG question answering benchmark, ECG-QA (PTB-XL version \cite{NEURIPS2023_d0b67349}), to test the model performance in terms of answering specific ECG screening questions. Specifically, we assessed the model across three groups of tasks: single-verify, single-choose, and single-query, each designed to probe different aspects of ECG interpretation. The expected answers from these questions are from a set of candidate templates, leading to this QA task as a multi-label classification task. Without loss of generality, we selected CSFM-Tiny for comparison. The results are presented in \figureautorefname{}~\ref{fig:grouped}f. This was compared with that of the Fusion Transformer model introduced in Oh \etal{}\cite{NEURIPS2023_d0b67349}, as well as with as well as with LLaVA (llama3-llava-next-8b version), a large language model capable of image-text querying, which serves as the baseline. Further details of the text prompting are provided in Supplementary Section S2.3. We reported macro-F1, which was computed per question by considering only the valid answer candidates for that question. As observed, The CSFM benefits from the pretraining, as observed from the superior performance against the Fusion Transformer structure. The baseline performance of LLaVA was unsurprisingly limited, likely due to its lack of domain-specific knowledge in ECG interpretation. More analysis results of ECG-QA are available in the next section.

Over the five scenarios examined, in certain cases, CSFM-Large, where applicable, occasionally exhibits slightly inferior performance compared to CSFM-Base. This observation suggests that its current dataset may be relatively insufficient to fully leverage the capacity of a larger model, in contrast to some existing large pretrained vision or language foundation models (e.g., GPT4), which benefit from extensive pretraining on vast datasets. Future research may investigate the scaling laws of training foundation models in the cardiac biosignal domain to optimize the balance between model capacity and available data.

\subsection*{CSFM generalizes across different ECG leads and ECG/PPG modalities}
Ideally, the model should be capable of learning distinctive representations regardless of the type of cardiac signal provided as input. To test this generalization capability, we evaluated the model performance using various channel configurations. % \figureautorefname{}~\ref{fig:channel}

\begin{figure}[tpb]
    \centering
    \includegraphics[width=0.31\linewidth]{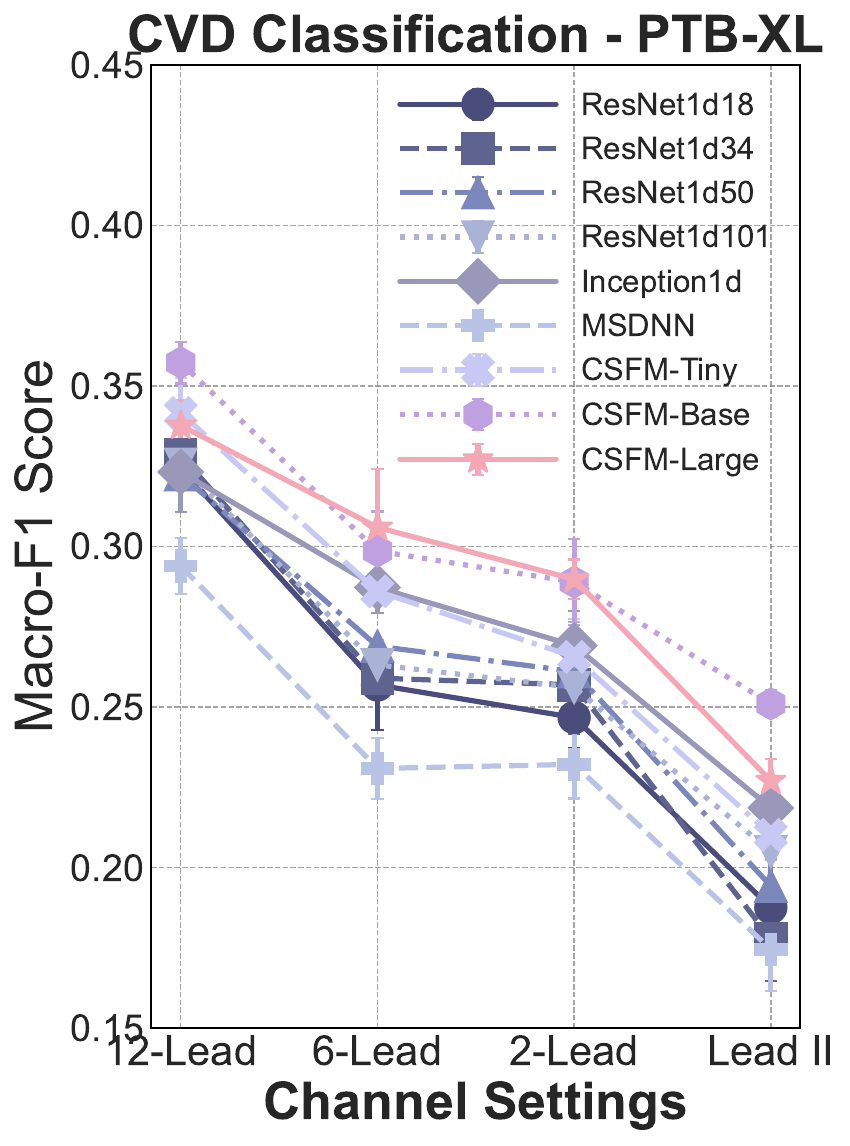}
    \includegraphics[width=0.31\linewidth]{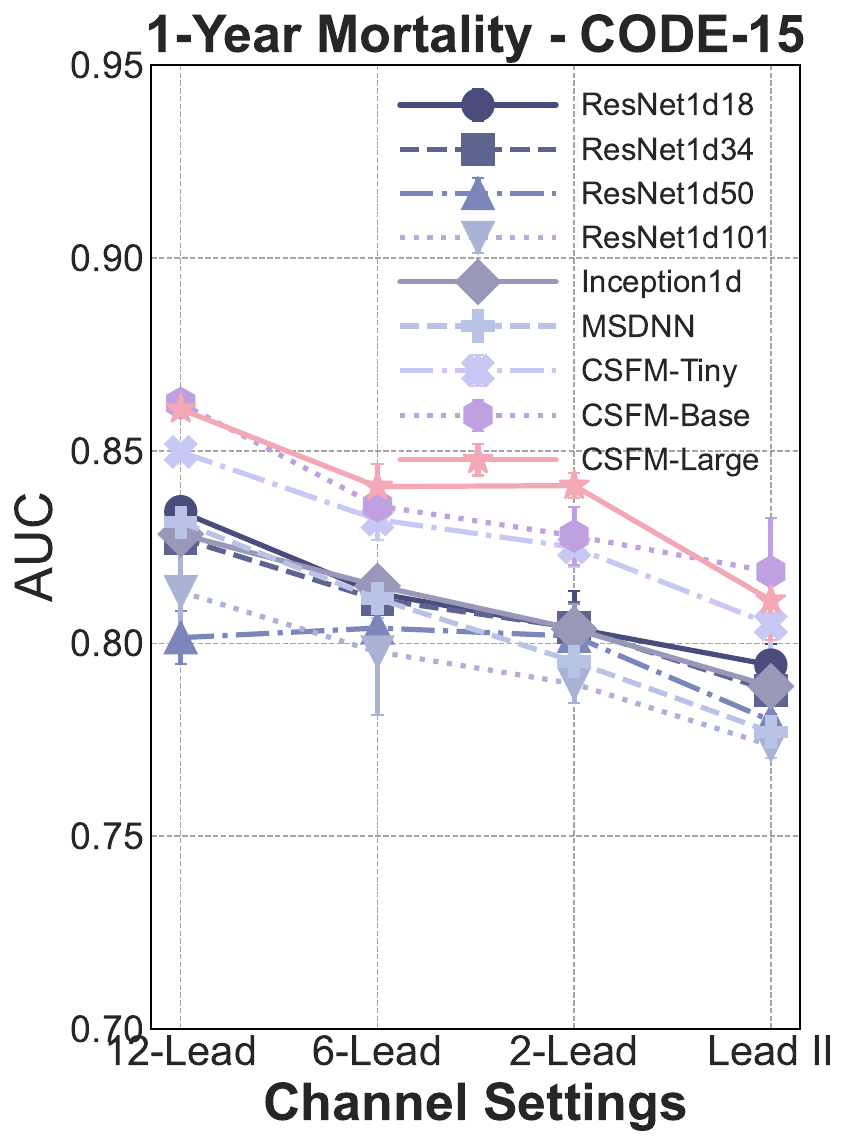}
    \includegraphics[width=0.31\linewidth]{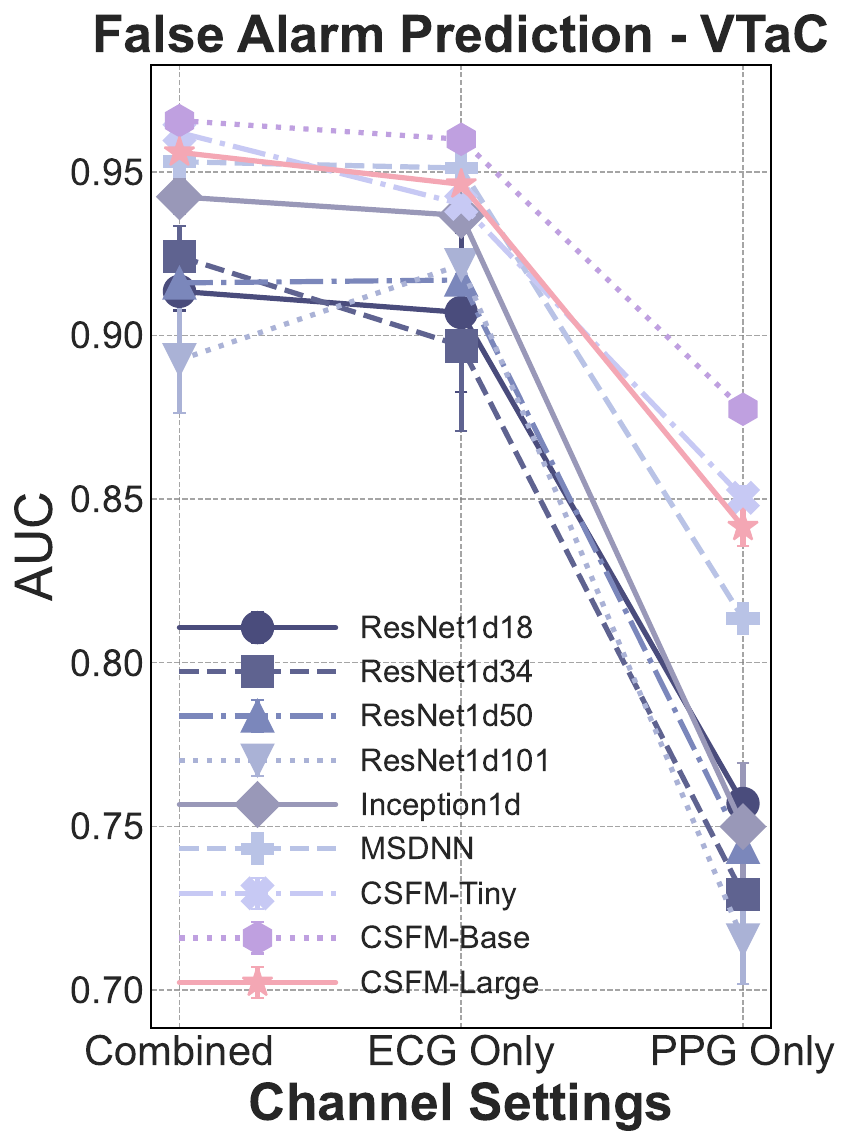}
    \caption{\textbf{Performance under different channel settings (full 12-Lead, 6-Lead, 2-Lead, Lead II) or the combinations of different sensing modalities (ECG and PPG).} In particular, 6-Lead utilizes \{Lead I, II, III, aVL, aVR, and aVF\}, and 2-Lead utilizes \{Lead II and V5\}. The experiments on ECG lead variations were performed for cardiovascular disease diagnosis on PTB-XL (leftmost), mortality prediction on CODE-15 (middle). In addition, we also examined the model's generalization performance across different sensing modalities for ICU false alarm prediction, on VTaC (rightmost).}
    \label{fig:channel}
\end{figure}

\noindent\textbf{Performance under Varied Lead Configurations of ECGs.} First, we assessed the transferability of our CSFM foundation model series across different ECG lead configurations. In many diagnostic environments, standard 12-lead ECGs may not be readily available or affordable \cite{reyna2022issues}, posing significant challenges for reusing models pretrained on specific lead configurations. Our foundation models, however, are designed to be adaptable across varied settings and demonstrate superior performance compared to conventional training methods. As shown in \figureautorefname{}~\ref{fig:channel}, experiments on the PTB-XL dataset (CVD diagnosis) and CODE-15 dataset (1-Year Mortality) confirm that our cardiac sensing foundation models outperform existing bespoke models across various lead configurations, including 12-lead, 6-lead (I, II, III, aVL, aVR, and aVF), 2-lead (II and V5), and single-lead (Lead II) setups.

\noindent\textbf{Performance under Varied ECG and PPG Settings.} Additionally, the availability of ECG and PPG signals can vary significantly in cardiac sensing applications. We examined our model's performance across scenarios where only ECG, only PPG, or a combination of both is available. This evaluation was conducted using VTaC-based false alarm prediction (as shown in the right side of \figureautorefname{}~\ref{fig:channel}) and VitalDB-based blood pressure reconstruction (as shown in \tableautorefname{}~\ref{tab:bp_table}). It is observed that our foundation models consistently demonstrated superior performance across these different signal modalities.

\noindent\textbf{Transfer from 12-Lead to Fewer-Lead Settings.} On the other hand, we conducted experiments to evaluate the transferability of models pretrained on 12-lead ECGs to settings with fewer leads. Specifically, we selected both conventional deep learning models and our cardiac sensing foundation models pretrained on PTB-XL (using 12-lead configurations) and subsequently fine-tuned them on PTB-XL subsets with 6, 2, and 1 leads. We assessed performance when fine-tuning with 100\%, 50\%, and 10\% of the full training set of PTB-XL. This was similar to the protocols proposed in one previous work \cite{kiyasseh2021clinical}. The results are presented in \tableautorefname{}~\ref{tab:transfer}. For conventional models, transferring pretrained weights to different lead configurations is not trivial because their architectures often include layers with input channel sizes fixed to the number of leads (e.g., the first 1D convolutional layer in the ResNet1d series is designed for 12 channels). To address this issue, we randomly reinitialized the weights of these input-specific layers during transfer learning, while transferring the remaining layers. By contrast, our foundation models are channel-agnostic, enabling direct transfer learning without the need to reinitialize input-specific layers. As observed in \tableautorefname{}~\ref{tab:transfer}, CSFMs consistently outperform conventional approaches. Notably, even when fine-tuning with only 10\% of the training data, our model achieves performance comparable to conventional models trained on 100\% of the dataset.

% \xg{to be added}
\noindent\textbf{Lead-Related ECG Question Answering with Only Lead-II as Input.} We also investigated whether CSFM can answer questions that typically depend on lead specifics (including keyword \textit{``lead''}), even when only a single lead (Lead II) is provided as input. This represents a particularly challenging task, as most clinically relevant spatial patterns in ECG interpretation require multiple leads for accurate assessment, especially when questions involve features observable in other leads. The goal is to explore whether patterns typically distributed across multiple leads can, to some extent, be inferred from Lead II alone. We compared the performance of CSFM and the Fusion Transformer under two input settings: full 12-lead ECG and Lead II only. Notably, CSFM with only Lead II input achieved performance comparable to that of the Fusion Transformer using the full 12-lead input. This suggests that CSFM's pretraining enables it to capture and uncover global information that generalizes beyond the visible lead.

\begin{table}[tpb]
    \centering
    \caption{\textbf{Blood pressure measurement prediction for both continuous waveforms (arterial blood pressure-ABP waveform) and numerical values (systolic blood pressure-SBP and diastolic blood pressure-DBP), under the unit of mmHg.} The performance of different methods was benchmarked under different settings (with ECG, PPG, or combined modalities as input). The continuous waveform was reconstructed with our foundation model with additional dense regression head or by compared sequence to sequence generation models. Their performance was measured by mean absolute error (MAE) and root mean square error (RMSE). Subsequently, we derived the maximal and minimal value from the continuous blood pressure waveform as predicted SBP and DBP numeric values. Best values are in bold, and second best are underlined.}
    \label{tab:bp_table}
    \resizebox{0.6\linewidth}{!}{%
    \begin{tabular}{lcc|cccccc}
    \toprule
    \multirow{2}{*}{\textbf{Methods}} & \multirow{2}{*}{\textbf{ECG}} & \multirow{2}{*}{\textbf{PPG}} & \multicolumn{2}{c}{\textbf{ABP} (waveform)} & \multicolumn{2}{c}{\textbf{SBP} (numeric)} & \multicolumn{2}{c}{\textbf{DBP} (numeric)} \\
    \cmidrule(lr){4-5} \cmidrule(lr){6-7} \cmidrule(lr){8-9}
     & & & MAE & RMSE & MAE & RMSE & MAE & RMSE \\
    \midrule
    % UNet group (reordered)
  % \rowcolor[HTML]{AAB2D6}  
       & \checkmark &  & 9.077  & 12.739 & 14.380 & 18.130 & 7.878  & 11.120 \\ % ECG only
    % \rowcolor[HTML]{AAB2D6}  
     UNet1D
  &  & \checkmark & 10.874 & 14.511 & 12.951 & 16.712 & 8.182  & 11.257 \\ % PPG only
    % \rowcolor[HTML]{AAB2D6}  
     & \checkmark & \checkmark & 7.803  & 10.927 & 12.509 & 16.056 & 7.722  & 10.876 \\ % Both sensors
       \arrayrulecolor[HTML]{AAB2D6}
  \midrule
% \rowcolor[HTML]{B9C3E6}
% \multirow{3}{*}{BiLSTM} 
       & \checkmark &  & 7.821  & 10.833 & 10.076 & 13.514 & 5.865  & 8.910 \\ % ECG only
  % \rowcolor[HTML]{B9C3E6}
 BiLSTM    &  & \checkmark & 9.420  & 12.412 & 12.086 & 15.819 & 7.593  & 10.505 \\ % PPG only
  % \rowcolor[HTML]{B9C3E6}
     & \checkmark & \checkmark & 8.439  & 11.201 & 10.864 & 14.220 & 6.991  & 9.919 \\ % Both sensors
    \midrule
% \rowcolor[HTML]{9998B8}
     % \multirow{3}{*}{AutoEncoder\cite{gu2020cross}} 
       & \checkmark &  & 7.804  & 11.064 & 9.440  & 12.833 & 5.780  & 8.974 \\ % ECG only
  % \rowcolor[HTML]{9998B8}
 AutoEncoder\cite{gu2020cross}    &  & \checkmark & 9.368  & 12.495 & 11.604 & 15.164 & 7.324  & 10.289 \\ % PPG only
 % \rowcolor[HTML]{9998B8}
      & \checkmark & \checkmark & 6.381  & 9.010  & 8.145  & 11.091 & 5.212  & 8.244 \\ % Both sensors
    \midrule
      \arrayrulecolor{black}

    \rowcolor[HTML]{EEEEFA}
    % \multirow{3}{*}{CSFM-Tiny} 
       & \checkmark &  & 5.052  & 7.283  & 6.919  & 9.923  & 4.027  & 7.093 \\ % ECG only
    \rowcolor[HTML]{EEEEFA}
    CSFM-Tiny  &  & \checkmark & 7.495  & 10.448 & 9.757  & 13.465 & 6.079  & 9.200 \\ % PPG only
       \rowcolor[HTML]{EEEEFA}
    & \checkmark & \checkmark & 3.281  & 4.783  & 4.533  & \underline{6.699}  & 2.812  & \underline{5.982} \\ % Both sensors
    % \midrule
    \rowcolor[HTML]{EAD4E3}  
       & \checkmark &  & 4.549  & 6.723  & 6.423  & 9.332  & 3.730  & 6.903 \\ % ECG only
     \rowcolor[HTML]{EAD4E3}  
    CSFM-Base
     &  & \checkmark & 7.179  & 10.148 & 9.403  & 13.163 & 5.803  & 8.974 \\ % PPG only
        \rowcolor[HTML]{EAD4E3}  
   & \checkmark & \checkmark & \underline{3.203}  & \underline{4.717}  & \underline{4.497}  & 6.748  & \underline{2.808}  & 5.990 \\ % Both sensors
    % \midrule
    \rowcolor[HTML]{F3C6D5}  
       & \checkmark &  & 4.404  & 6.568  & 6.402  & 9.202  & 3.025  & 6.320 \\ % ECG only
       \rowcolor[HTML]{F3C6D5}  
    CSFM-Large
   &  & \checkmark & 6.988  & 9.990  & 9.104  & 12.875 & 5.103  & 8.723 \\ % PPG only
        \rowcolor[HTML]{F3C6D5}  
   & \checkmark & \checkmark & \textbf{3.102}  & \textbf{4.690}  & \textbf{4.420}  & \textbf{6.354}  & \textbf{2.753}  & \textbf{5.901} \\ % Both sensors
    \bottomrule
    \end{tabular}%
    }
\end{table}

\begin{figure}
    \centering
    \includegraphics[width=0.8\linewidth]{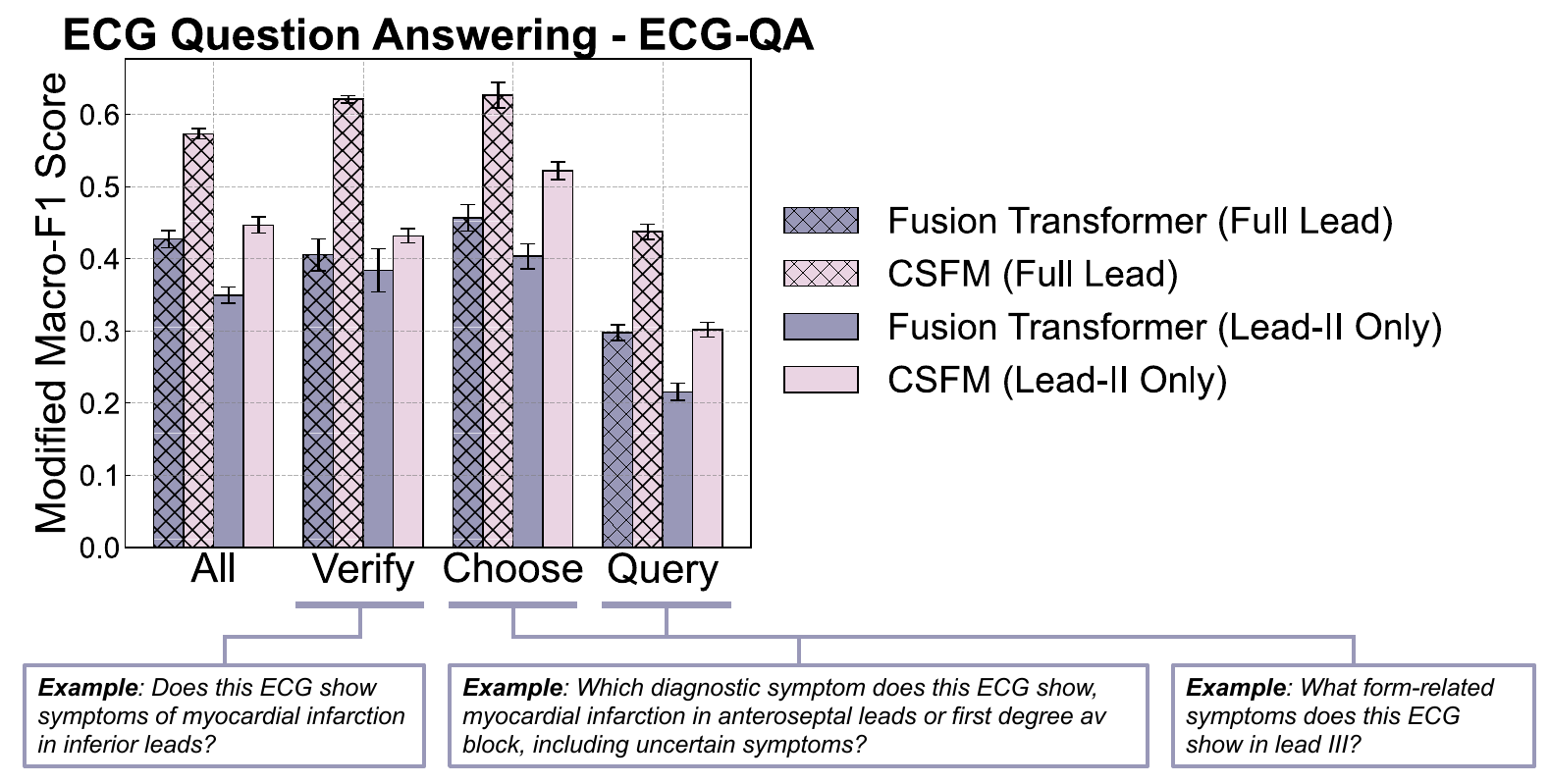}
    \caption{\textbf{ECG Question Answering, for lead-related questions with only lead-II ECG as input.} We selected a subset of questions that are intuitively related to leads (including the keyword \textit{``lead''}), with representative examples illustrated in the accompanying plot. We compared the performance of the Fusion Transformer and CSFM when restricted to Lead II input, and also reported their performance when trained or fine-tuned on all 12 leads. Performance was measured using the macro-F1 score, calculated based on the valid candidate options for each question.}
    \label{fig:ecg-qa}
\end{figure}

\begin{table}[tbph!]
    \centering
    \caption{\textbf{Results of transfer learning settings of CSFM.} We initialized our model with weights fully fine tuned on the 12-lead ECGs from the PTB-XL dataset, and then further fine tuned it on configurations with fewer leads (i.e., 6-lead, 2-lead, and single-lead Lead II). Training was conducted on 100\%, 50\%, or 10\% of the train subset, respectively, and the performance of the test subset was measured using the Macro-F1 score. In addition, we report the performance gap (rectangle alongside each number, {\improve{}} indicates improved and {\degrade{}} degraded) compared to models directly trained (conventional deep learning based on train from scratch, and CSFM based on finetuning) on the corresponding lead configurations with 100\% train subset. Best values are in bold, and second best are underlined.} 
    \vspace{-5pt}
    \label{tab:transfer}
    \resizebox{1.\linewidth}{!}{
    \begin{tabular}{l|ccccccccc}
      \toprule
      \multirow{2}{*}{\textbf{Methods}}   &  \multicolumn{3}{c}{\textbf{6-Lead}} & \multicolumn{3}{c}{\textbf{2-Lead}} & \multicolumn{3}{c}{\textbf{Lead II}} \\
      \cmidrule(lr){2-4} \cmidrule(lr){5-7} \cmidrule(lr){8-10}
      & 100\% & 50\% & 10\%  & 100\% & 50\% & 10\% & 100\% & 50\% & 10\%  \\
      \midrule
        ResNet1d18    & 0.290 \improve{+0.052}
        & 0.279 \improve+{0.041}
        & 0.228 \degrade{-0.010}
        & \underline{0.272} \improve{+0.029}
        & 0.252 \improve{+0.009}
        & 0.184 \degrade{-0.059}
        & 0.198 \improve{+0.020}
        & 0.175 \degrade{-0.003}
        & 0.139 \degrade{-0.039} \\
        ResNet1d34 & 
        0.264 \degrade{-0.002} & 
        {0.260 \degrade{-0.006}} & 
        {0.216 \degrade{-0.050}} & 
        {0.271 \improve{+0.032}} & 
        {0.249 \improve{+0.010}} & 
        {0.167  \degrade{-0.072}} & 
        {0.202  \improve{+0.020}} & 
        {\underline{0.189}  \improve{+0.007}} & 
        {0.144 \degrade{-0.038}} \\
        ResNet1d50 & 0.270 \degrade{-0.000} & 0.264 \degrade{-0.006} & 0.217 \degrade{-0.053} & 0.275 \improve{+0.003} & 0.258 \degrade{-0.014} & 0.196 \degrade{-0.076} & 0.195 \improve{+0.008} & 0.185 \degrade{-0.002} & 0.139 \degrade{-0.048} \\
        ResNet1d101 & 0.236 \degrade{-0.037} & 0.238 \degrade{-0.035} & 0.182 \degrade{-0.091} & 0.216 \degrade{-0.021} & 0.206 \degrade{-0.031} & 0.143 \degrade{-0.094} & 0.170 \degrade{-0.039} & 0.151 \degrade{-0.058} & 0.116 \degrade{-0.093} \\
        Inception1d \cite{ismail2020inceptiontime} & 0.270 \degrade{-0.007} & 0.233 \degrade{-0.044} & 0.182 \degrade{-0.095} & 0.253 \degrade{-0.003} & 0.240 \degrade{-0.016} & 0.168 \degrade{-0.088} & 0.181 \degrade{-0.042} & 0.178 \degrade{-0.045} & 0.145 \degrade{-0.078} \\
        MSDNN \cite{lai2023practical} & 0.262 \improve{+0.024} & 0.249 \improve{+0.011} & 0.196 \degrade{-0.042} & 0.259 \improve{+0.040} & 0.244 \improve{+0.025} & 0.189 \degrade{-0.030} & 0.183 \improve{+0.018} & 0.180 \improve{+0.015} & 0.147 \degrade{-0.018} \\
        % \midrule
        \rowcolor[HTML]{EEEEFA}
        CSFM-Tiny & 0.268 \degrade{-0.019} & 0.258 \degrade{-0.029} & 0.248 \degrade{-0.039} & 0.267 \degrade{-0.013} & 0.261 \degrade{-0.019} & 0.216 \degrade{-0.064} & 0.206 \degrade{-0.003} & \underline{0.189} \degrade{-0.020} & \underline{0.167} \degrade{-0.042} \\
        \rowcolor[HTML]{EAD4E3}
        CSFM-Base & \textbf{0.312} \degrade{-0.034} & \textbf{0.307} \degrade{-0.039} & \textbf{0.272} \degrade{-0.074} & \textbf{0.291} \degrade{-0.017} & \textbf{0.280} \degrade{-0.028} & \underline{0.237} \degrade{-0.071} & \underline{0.215} \degrade{-0.033} & 0.183 \degrade{-0.065} & 0.165 \degrade{-0.083} \\
        \rowcolor[HTML]{F3C6D5}
        CSFM-Large & \underline{0.310} \degrade{-0.007} & \underline{0.296} \degrade{-0.021} & \underline{0.261} \degrade{-0.056} & \textbf{0.291} \improve{+0.010} & \underline{0.279} \degrade{-0.003} & \textbf{0.251} \degrade{-0.031} & \textbf{0.226} \improve{+0.008} & \textbf{0.209} \degrade{-0.009} & \textbf{0.182} \degrade{-0.036} \\
        \bottomrule
    \end{tabular}}
\end{table}

\begin{figure}[htpb!]
    \centering
    \includegraphics[width=0.8\linewidth]{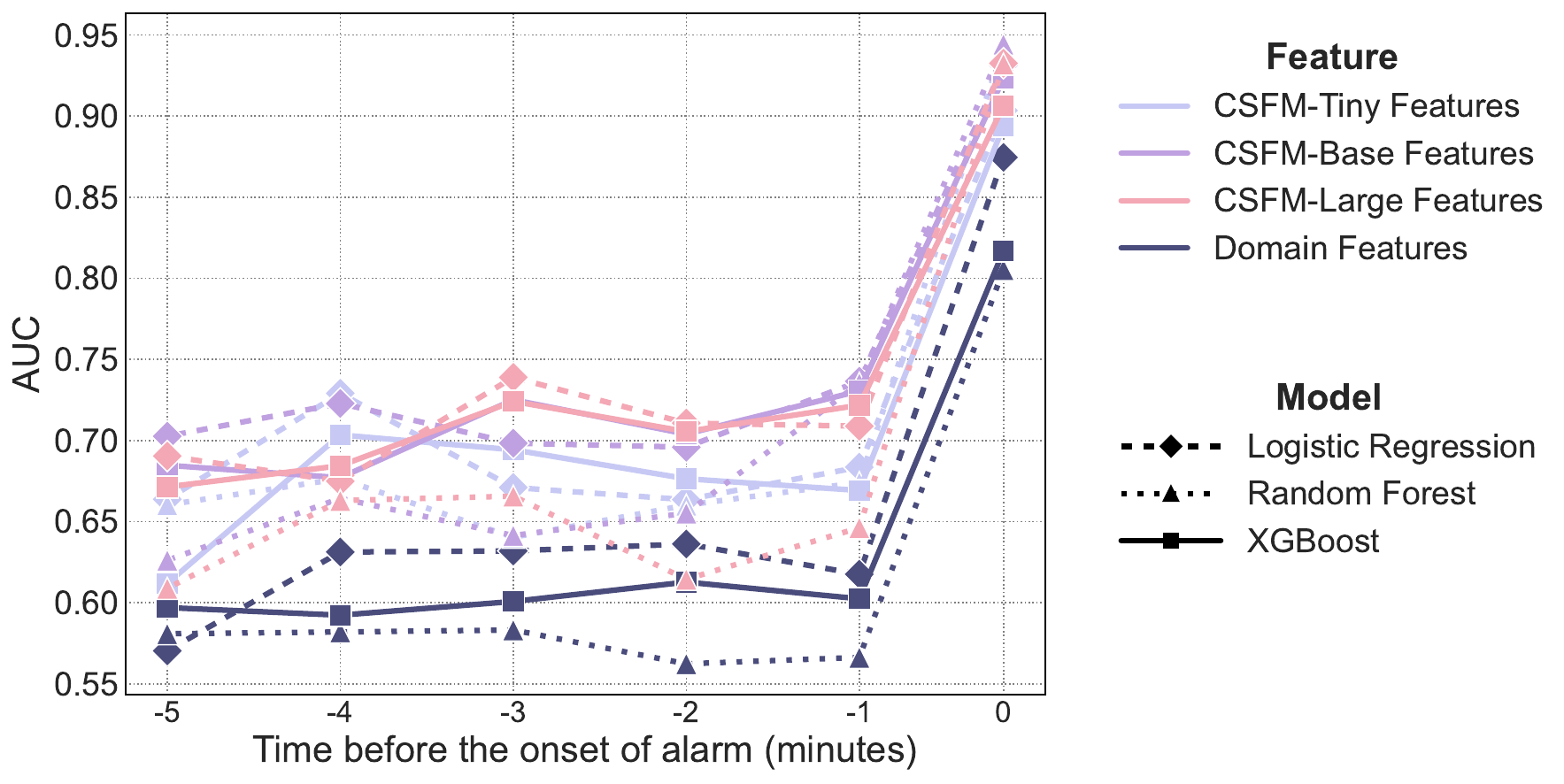}
    \caption{\textbf{Comparison results of the predictive performance between domain features (hand-crafted features) and the features extracted from our cardiac sensing foundation models, on VTaC.} We compared the predictive performance of signals collected from different timestamps, including 5,4,3,2,1 or 0 minute prior to the onset of ICU alarm. Each feature set was evaluated using three classical machine learning classifiers: Logistic Regression (dashed line), Random Forest (dotted line), and XGBoost (solid line).}
    \label{fig:vtac_time}
\end{figure}

\begin{figure}[htpb!]
    \centering
    \includegraphics[width=0.265\linewidth]{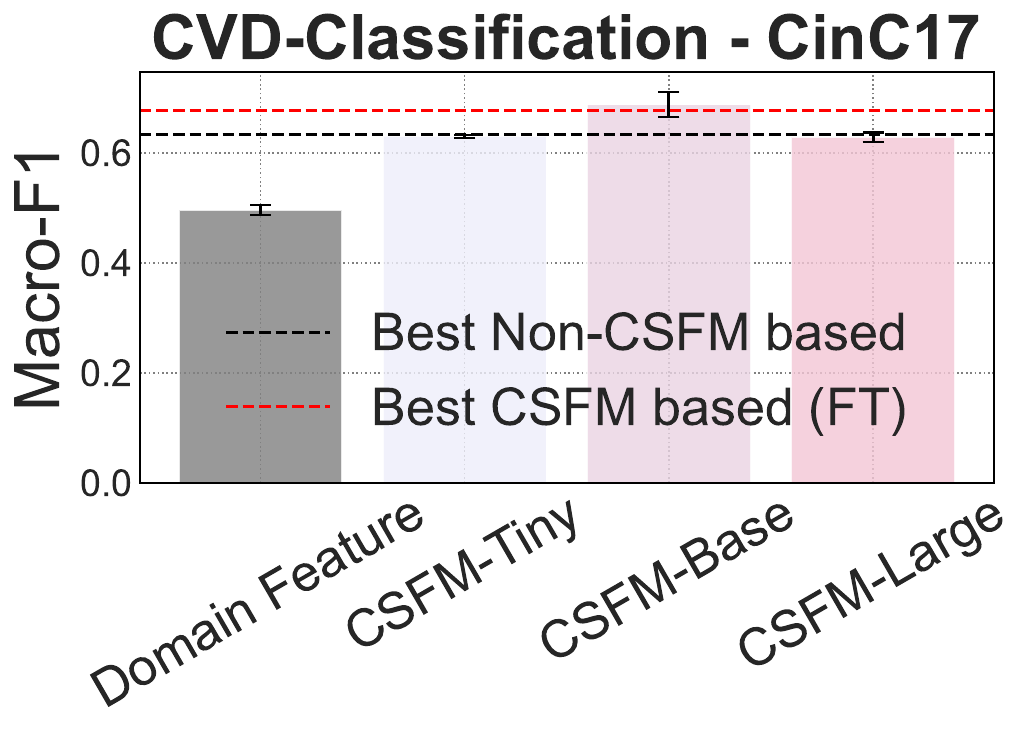}
    \includegraphics[width=0.34\linewidth]{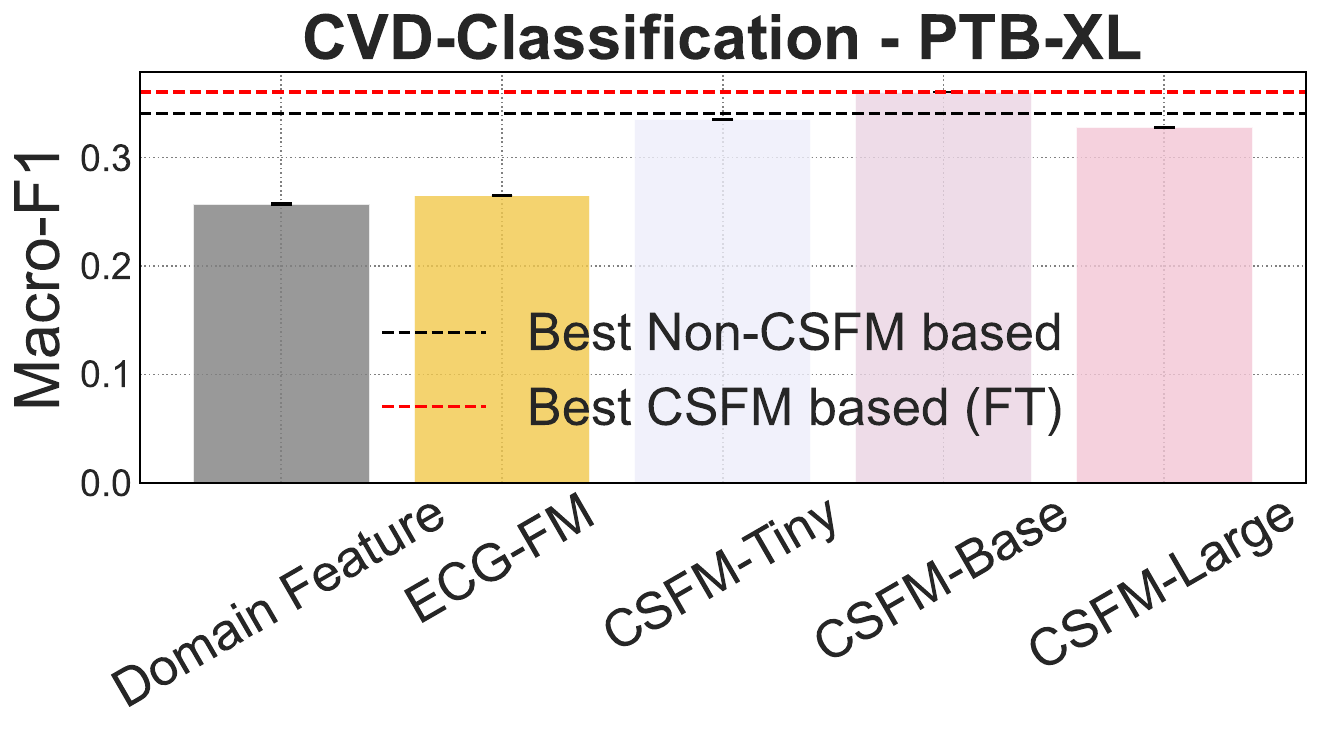}
    \includegraphics[width=0.34\linewidth]{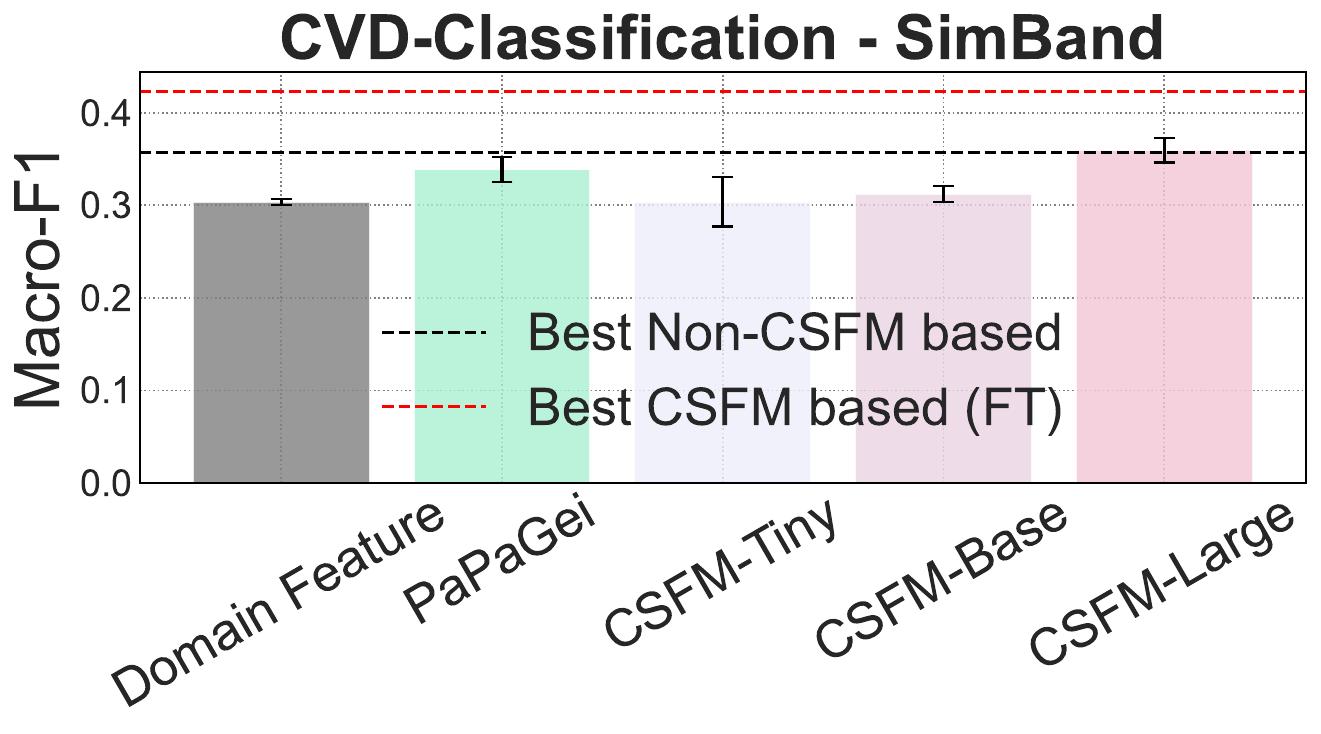}
    \caption{\textbf{Comparisons of domain features, open-sourced foundation model extracted features, and our foundation model extracted features for cardiovascular disease classification.}  Domain features were extracted using open-source toolkits such as NeuroKit2 for ECG and pyPPG for PPG, while open-source foundation models include ECG-FM and PaPaGei. All extracted features were used as input to an XGBoost classifier for downstream classification. For reference, the figure also includes the end-to-end performance of the best CSFM-based method (via fine tuning) and the best non-CSFM baseline.}
    \label{fig:domain_feature}
\end{figure}

\subsection*{CSFM acts as an effective feature extractor}

Analysis of biomedical signals, particularly ECG and PPG, has advanced considerably over the years. Manual extraction of commonly-used domain-specific features is still a popular approach. However, achieving robustness and generalizability across diverse data collection protocols and devices remains a significant challenge. Limited efforts have been made to develop a unified toolbox for extracting useful features from such heterogeneous settings, whether through traditional manual feature engineering or deep neural networks. We demonstrate that the representations learned by our pretrained models serve as effective embeddings, accommodating these variations and enhancing performance across different settings.

\noindent\textbf{Comparison to Manually Engineered Features.} First, we assessed the features extracted by the CSFM series by comparing them against domain-knowledge-driven bespoke features. To achieve this, we leveraged established biomedical signal processing toolboxes to extract relevant features from each modality separately. Specifically, we used NeuroKit2\footnote{\url{https://neuropsychology.github.io/NeuroKit/}} and pyPPG\footnote{\url{https://pyppg.readthedocs.io/}} to extract features from ECG and PPG signals, respectively, encompassing temporal, frequency, and morphological dimensions. It should be noted that feature vectors were extracted from each ECG lead individually; when multiple leads were available, average values across leads were computed. Further details regarding these features are provided in the Supplementary Section S2.1.

\noindent\textbf{Predictive Performance Over Time Horizons.} In addition to diagnostic tasks, we benchmarked predictive performance for recognizing ICU false alarms over various time horizons. Specifically, we extracted 10-second recordings at multiple intervals preceding the onset of alarms—immediately before the alarm, as well as 1, 2, 3, 4, and 5 minutes prior. We then extracted both domain-specific features and model-derived embeddings, and applied three classifiers (logistic regression, random forest, and XGBoost) to evaluate performance across these time intervals. The results are displayed in \figureautorefname~\ref{fig:vtac_time}. Our findings indicate that the model-derived embeddings consistently outperform the domain-specific features, and among the classifiers, XGBoost generally achieves the best performance for both feature sets.

\noindent\textbf{Comparison to Features Extracted from State-of-the-Art Time Series Foundation Models.} Furthermore, we compared the performance of CSFM-derived embeddings against those extracted from general time series models and from dedicated ECG/PPG foundation models. Specifically, we benchmark our embeddings against those obtained from ECG-FM \cite{mckeen2024ecg} (compatible with 12-lead ECG) and PaPaGei \cite{pillai2024papagei} (compatible with PPG), with detailed settings described in Supplementary Section S2.2. Their performance was assessed on PTB-XL and SimBand, separately. All these features are trained with a XGBoost classifier. As shown in \figureautorefname{}~\ref{fig:domain_feature}, our foundation model outperforms the alternatives. It is also noteworthy that, in some cases, the embeddings extracted from our foundation models—when used in conjunction with an XGBoost classifier—yield performance comparable to that of fully fine-tuned foundation models, and conventional models trained from scratch. This demonstrates the viability of directly employing our foundation models as generic feature extractors for diagnostic applications.

\subsection*{CSFM facilitates cross-modality reconstruction and augmentation}

Due to the limitations of advanced sensing solutions, especially in some resource-limited scenarios, e.g., LMICs, collecting standard 12-Lead ECG is often challenging. This motivates our investigation into two specific applications to evaluate the versatility of our foundation models. Similar to the \hyperlink{\ref{sec:task}}{Vital Sign Measurement}, we added a dense regression module on top of the transformer module to generate dense outputs. 
% 1) Generating ECG from PPG 2) Reconstructing 12-Lead ECG from Single-Lead ECG. 

\noindent\textbf{From PPG to ECG.} 
In this setting, we generated ECG waveforms from PPG signals and evaluated our pretrained model's performance on atrial fibrillation (AF) detection. Table~\ref{tab:ppg2ecg} summarizes the results, including both waveform reconstruction performance on the held-out test set of VitalDB and the transfer performance between synthetic ECGs generated from the SimBand dataset and real ECGs from CinC17. Specifically, we trained our model on VitalDB and then applied the trained model to the original SimBand dataset (selecting only Normal and AF cases) to generate synthetic Lead-II ECG waveforms. To comprehensively evaluate the quality of these generated ECG waveforms, we assessed the transfer performance between the synthetic SimBand-ECG and CinC17, reporting performance metrics in terms of F1 score and AUC.

\noindent\textbf{From Single-Lead ECG to 12-Lead ECG.}
Here, we reconstructed the full 12-lead ECGs from single-lead data, as synthetic data. The reconstruction model was trained on MIMIC-IV (using Lead-II ECG to generate a full 12-lead ECG) and subsequently applied to the PTB-XL dataset to produce synthetic ECG recordings. We assessed the quality of these reconstructions under both train-real/test-synthetic and train-synthetic/test-real settings, with performance metrics detailed in Table~\ref{tab:ecg_aug}.

Across these two tasks, the reconstructed data generated by our foundation models demonstrates superior performance compared to the original data. However, a noticeable gap between real and synthetic data persists, as evidenced by the discrepancies observed between train-real/test-synthetic and train-synthetic/test-real evaluations. Future work could explore conditional generative training or diffusion models to enhance the plausibility and fidelity of the generated signals.

% \xg{discuss limitations, like conditional generation, or diffusion models, can add to the plausibility of the model.}

\begin{table}[tbp]
\centering
\caption{\textbf{Cross-modality reconstruction and augmentation results.} \textbf{a.} PPG to ECG Reconstruction. (1) The reconstruction was performed on VitalDB, and the waveform reconstruction performance was reported on the held-out test set of VitalDB. (2) Subsequently, we applied the adapted model to the original SimBand dataset to generate synthetic Lead-II ECG waveforms. To comprehensively test the quality of generated ECG waveforms, we conducted two experimental settings: train on synthetic ECG from SimBand (normal versus AF), and test on real ECG on CinC17 (normal versus AF), and vice versa. The performance is reported using F1 and AUC. Best values are in bold, and second best are underlined. \textbf{b.} Single-lead ECG to 12-lead ECG Augmentation.
(1) Likewise, we leveraged MIMIC-IV (training set) to perform reconstruction of Lead-II ECG to full 12-Lead ECG. Subsequently, we applied the trained model on PTB-XL to generate synthetic ECG recordings. The reconstruction performance is measured within the whole set of PTB-XL.
(2)  Based on the synthetic recordings, we performed both train-real test-synthetic and train-synthetic test-real settings, to assess the quality of generated ECG signals. The performance is reported using F1 and AUC. Best values are in bold, and second best are underlined.}
\vspace{-5pt}
\makebox[0pt][l]{\raisebox{2cm}{\hspace{-0cm}\Large{\textbf{a}}}}%
\begin{minipage}[b]{0.49\textwidth}
\centering

\vspace{-10pt}
\label{tab:ppg2ecg}
\bigskip
\resizebox{\linewidth}{!}{
\begin{tabular}{l|cccccc}
\toprule
\multirow{2}{*}{\textbf{Methods}}  & \multicolumn{2}{c}{\shortstack{\textbf{Waveform} \\ \textbf{Reconstruction}}} & \multicolumn{2}{c}{\shortstack{\textbf{Train-Real} \\ \textbf{Test-Synthetic}}} & \multicolumn{2}{c}{\shortstack{\textbf{Train-Synthetic} \\ \textbf{Test-Real}}} \\
\cmidrule(lr){2-3}\cmidrule(lr){4-5}\cmidrule(lr){6-7}
& MAE & RMSE & F1 & AUC & F1 & AUC \\ 
\midrule 
UNet1d         & 0.608 & 0.964 & 0.427 & 0.591 & 0.276 & 0.558 \\
BiLSTM       & 0.607 & 0.953 & 0.536 & 0.648 & 0.340 & 0.622 \\
AutoEncoder\cite{gu2020cross}  & 0.585 & 0.927 & 0.442 & 0.784 & 0.353 & 0.600\\
CSFM-Tiny    & 0.532 & 0.863 & \underline{0.690} & \underline{0.812} & \best{0.365} & 0.669 \\
CSFM-Base    & \underline{0.524} & \underline{0.852} & 0.632 & 0.815 & \underline{0.364} & \best{0.690} \\
CSFM-Large   & \textbf{0.516} & \best{0.840} & \best{0.692} & \best{0.820} & 0.353 & \underline{0.688} \\
\bottomrule
\end{tabular}}
\end{minipage}\hfill
\makebox[0pt][l]{\raisebox{2cm}
{\hspace{0cm}\Large{\textbf{b}}}}%
\begin{minipage}[b]{0.49\textwidth}
\centering

\vspace{-10pt}
\label{tab:ecg_aug}
\bigskip
\resizebox{\linewidth}{!}{
\begin{tabular}{l|cccccc}
\toprule
\multirow{2}{*}{\textbf{Methods}}  & \multicolumn{2}{c}{\shortstack{\textbf{Waveform} \\ \textbf{Reconstruction}}} & \multicolumn{2}{c}{\shortstack{\textbf{Train-Real} \\ \textbf{Test-Synthetic}}} & \multicolumn{2}{c}{\shortstack{\textbf{Train-Synthetic} \\ \textbf{Test-Real}}} \\
\cmidrule(lr){2-3}\cmidrule(lr){4-5}\cmidrule(lr){6-7}
& MAE & RMSE & F1 & AUC & F1 & AUC \\ 
\midrule 
UNet1d        &  0.530 & 0.907 & 0.120 & 0.758 & 0.070 & 0.567 \\
BiLSTM       &  0.543 & 0.906 & 0.094 & 0.761 & 0.031 & 0.514 \\
AutoEncoder\cite{gu2020cross}  &  0.533 & 0.904 & 0.107 & 0.754 &  0.047 & 0.565 \\
CSFM-Tiny    &  0.525 & \underline{0.903} & 0.123 & 0.785 & 0.101 & 0.731 \\
CSFM-Base    &  \underline{0.520} & 0.904 & \underline{0.134} & \textbf{0.793} & \underline{0.115} & \underline{0.736} \\
CSFM-Large   &  \textbf{0.510} & \textbf{0.898} & \textbf{0.158} & \underline{0.789} &  \textbf{0.139} & \textbf{0.755} \\
\bottomrule
\end{tabular}}
\end{minipage}
\end{table}

\begin{figure}[htpb]
    \centering
    \includegraphics[width=0.9\linewidth]{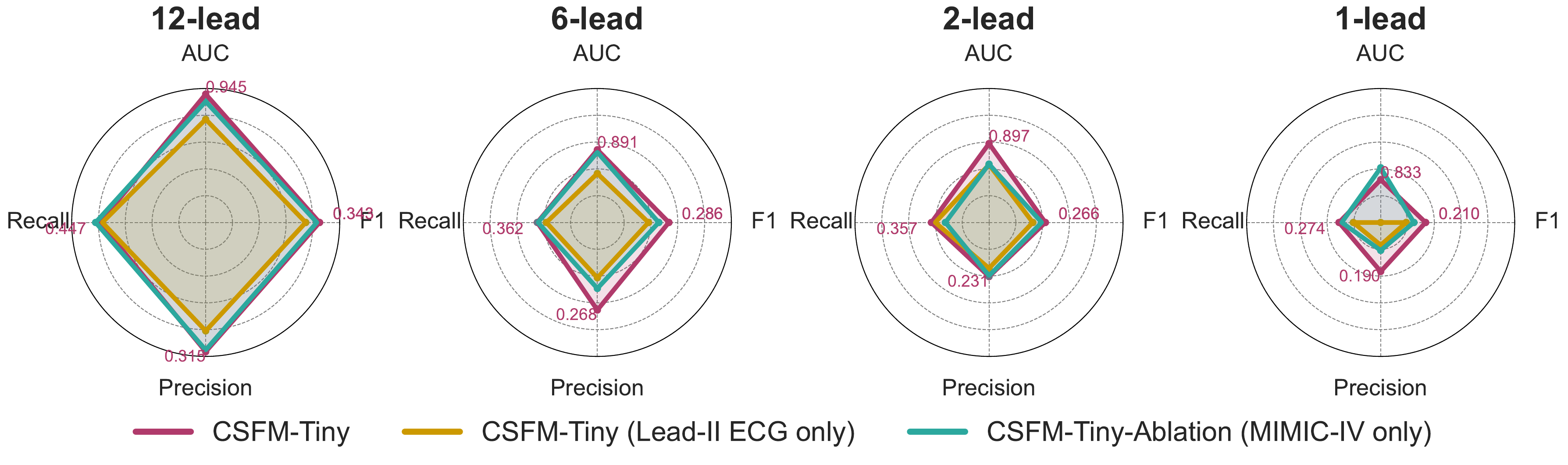}
    \caption{\textbf{Ablation study for different pretraining settings.} We compared our training strategies to other ``straightforward'' solutions to handling heterogeneous health records, (i) keeping the common channels only, i.e.,  Lead-II ECG, (ii) keeping one dataset only, e.g., MIMIC-IV including both 12-Lead ECGs and texts. Based on our training strategies and these two compared strategies, we assessed the performance disparity across varied lead settings on PTB-XL datasets. The radar axes in the figure are log-normalized and use the same range, for better visualization. } %\xg{A legend to illustrate different ablated versions}}
    \label{fig:ablation}
\end{figure}

\subsection*{Unification of heterogeneous datasets facilitates better pretraining}

The pretraining of our foundation model leverages vast amounts of heterogeneous health records aggregated from multiple datasets. To highlight the importance of integrating diverse data sources and modalities, we conducted comparative experiments using two alternative pretraining strategies: one utilizing only the MIMIC-IV-ECG dataset and the other restricting the data to common channels (specifically, Lead-II ECG only). These represent ``straightforward'' solutions when dealing with heterogeneous datasets, either by relying on just a single relatively large dataset or by selecting only the overlapping modalities. 

We benchmarked performance across different lead configurations on PTB-XL for these two ablated versions. As shown in \figureautorefname~\ref{fig:ablation}, our foundation model outperforms both alternative training strategies in most cases. Notably, in 1-lead settings, the model pretrained on MIMIC-IV (which includes both texts and 12-lead ECGs) outperforms the model trained on Lead-II ECG only (which aggregates Lead-II ECGs across datasets). This finding suggests that integrating data from multiple modalities can yield better performance, even when it compromises overall dataset size.

%\newpage
\section*{Discussion}

Our results demonstrate that the proposed cardiac sensing foundation model (CSFM) robustly generalizes across a wide range of clinical scenarios, devices, and input configurations. By integrating heterogeneous data sources, including ECGs, PPGs, and associated textual reports, the CSFM addresses the inherent fragmentation of traditional approaches, which are constrained by modality-specific silos and narrow task optimization. The transformer-based architecture, combined with a masked pretraining strategy, enables CSFM to learn rich, generalized representations that can be effectively fine-tuned for diverse downstream tasks, such as demographic information analysis, cardiovascular disease diagnosis, vital sign measurement, clinical outcome prediction, and ECG question answering. Extensive evaluations across multiple datasets (e.g., PTB-XL, CinC17, VitalDB, and CODE-15) confirm that our model consistently outperforms conventional deep learning models and bespoke feature-based methods. Notably, the model-derived embeddings not only enhance diagnostic accuracy and predictive performance but also exhibit exceptional transferability across various lead configurations and sensing modalities.

Despite these promising results, our work has several limitations. 
First, the interpretability of deep transformer models remains a challenge. Although our model captures intricate dependencies in cardiac biosignals, the ``black box'' nature of its internal representations can limit clinical trust and adoption. Second, while we utilized the embeddings from existing language models, further integration with large language models (LLMs) and large multimodal models (LMMs) with state-of-the-art methods, such as instruction tuning, may offer further improvements in interpretability and reasoning. Finally, the computational cost associated with training and deploying large-scale transformer architectures is non-trivial, potentially limiting accessibility in resource-constrained settings. Future research should focus on enhancing model interpretability, exploring hybrid strategies that directly incorporate LLMs/LMMs, and developing more computationally efficient training strategies.

\section*{Conclusion}

In conclusion, the cardiac sensing foundation model - CSFM represents an innovative advancement in the analysis of heterogeneous cardiac biosignals. By leveraging advanced transformer architectures and a generative pretraining strategy on large-scale, diverse datasets, our model learns robust, generalized representations that enhance diagnostic accuracy, predictive performance, and transferability across varied sensor configurations and clinical scenarios.

Our comprehensive evaluations demonstrate that CSFM consistently outperforms traditional, modality-specific methods, offering a scalable solution adaptable to both resource-rich and resource-constrained settings. While some challenges such as interpretability and computational cost remain, our findings underscore the potential of CSFM to transform cardiac monitoring and risk stratification. Overall, this work lays the groundwork for a new generation of versatile cardiac monitoring tools poised to improve patient care and outcomes in cardiovascular medicine. 

\section*{Data Availability}
The pretraining datasets MIMIC-III-WDB is available online\footnote{ \url{https://physionet.org/content/mimic3wdb-matched/1.0/}}, and their extensive clinical information (including subject-matched ECG reports) is available subject to corresponding data usage agreement\footnote{\url{https://physionet.org/content/mimiciii/1.4/}}. MIMIC-IV-ECG dataset is available online\footnote{\url{https://physionet.org/content/mimic-iv-ecg/1.0/}} as well. 
For the access of CODE-Full, please contact co-authors, Antonio H. Ribeiro and Antonio Luiz P. Ribeiro for more details. 

Regarding downstream validation datasets, VitalDB\footnote{\url{https://github.com/pulselabteam/PulseDB}} (preprocessed by PulseDB), CODE-15\footnote{\url{https://paperswithcode.com/dataset/code-15}}, VTaC \footnote{\url{https://www.physionet.org/content/vtac/1.0/}}, PTB-XL\footnote{\url{https://physionet.org/content/ptb-xl/1.0.3/}}, and CinC17\footnote{\url{https://physionet.org/content/challenge-2017/1.0.0/}} are all available online. Additionally, SimBand\footnote{\url{https://www.synapse.org/Synapse:syn23565056/wiki/608635}} are available based on access application, whilst ECG-QA\footnote{\url{https://github.com/Jwoo5/ecg-qa/tree/master/ecgqa/ptbxl}} (PTB-XL version) is available online with associated processing scripts.

\section*{Code Availability}
The pretrained model weights and the inference scripts will be made available \url{https://github.com/guxiao0822/Cardiac-Sensing-FM}. 

\section*{Acknowledgments}

D.A.C. was supported by the Pandemic Sciences Institute at the University of Oxford; the National Institute for Health Research (NIHR) Oxford Biomedical Research Centre (BRC); an NIHR Research Professorship; a Royal Academy of Engineering Research Chair; the Wellcome Trust funded VITAL project (grant 204904/Z/16/Z); the EPSRC (grant EP/W031744/1); and the InnoHK Hong Kong Centre for Cerebro-cardiovascular Engineering (COCHE). 

\section*{Author Contributions}
D.A.C. conceived and supervised the project, and revised the manuscript. X.G. conceived and designed the study, curated data, conducted experiments and data analysis, and drafted the manuscript. W.T. conducted experiments and revised the manuscript. J.H. performed the data analysis and revised the manuscript. Z.L. curated data, conducted experiments, and revised the manuscript. All the other authors significantly contributed to methodology design, result interpretation, and manuscript revision and finalization. 

\section*{Competing Interests}
The authors declare no competing interests.

\newpage
\bibliography{ref}

% \section*{Author contributions statement}
% \input{additional}
\end{document}